\def\year{2018}\relax
\documentclass[letterpaper]{article} 
\usepackage{aaai18}  
\usepackage{times}  
\usepackage{helvet}  
\usepackage{courier}  
\usepackage{url}  
\usepackage{graphicx}  

\usepackage[colorlinks=true,linkcolor=blue,urlcolor=blue,citecolor=blue]{hyperref}
\usepackage{amsmath}\allowdisplaybreaks
\usepackage{xspace,amssymb}
\usepackage{dsfont}
\usepackage{algorithm}
\usepackage{algorithmic}
\usepackage{subfigure}
\usepackage[capitalize,noabbrev]{cleveref}

\newcommand{\EE}{\mathbb{E}}
\newcommand{\PP}{\mathbb{P}}
\newcommand{\RR}{\mathbb{R}}
\newcommand{\bA}{\boldsymbol{A}}
\newcommand{\bC}{\boldsymbol{C}}

\newcommand{\bK}{\boldsymbol{K}}
\newcommand{\bM}{\boldsymbol{M}}

\newcommand{\bS}{\boldsymbol{S}}
\newcommand{\bU}{\boldsymbol{U}}
\newcommand{\bX}{\boldsymbol{X}}
\newcommand{\bY}{\boldsymbol{Y}}

\newcommand{\bb}{\boldsymbol{b}}

\newcommand{\bx}{\boldsymbol{x}}
\newcommand{\by}{\boldsymbol{y}}
\newcommand{\bbeta}{\boldsymbol{\beta}}
\newcommand{\btheta}{\boldsymbol{\theta}}

\newcommand{\bOne}{\boldsymbol{1}}
\newcommand{\cB}{\mathcal{B}}

\newcommand{\cE}{\mathcal{E}}
\newcommand{\cF}{\mathcal{F}}
\newcommand{\cG}{\mathcal{G}}
\newcommand{\cH}{\mathcal{H}}
\newcommand{\cS}{\mathcal{S}}
\newcommand{\argmax}{\mathrm{argmax}}
\newcommand{\norm}[1]{\left\| #1 \right\|}
\newcommand{\abs}[1]{\left| #1 \right|}
\newcommand{\trace}{\mathrm{trace}}
\newtheorem{theorem}{Theorem}
\newtheorem{corollary}[theorem]{Corollary}
\newtheorem{lemma}[theorem]{Lemma}
\newenvironment{proof}{\noindent {\textbf{Proof. }}}{$\Box$ \medskip}

\newcommand\numberthis{\addtocounter{equation}{1}\tag{\theequation}}
\begin{document}
%
\title{Online Clustering of Contextual Cascading Bandits}
\author{Shuai Li\\
The Chinese University of Hong Kong\\
shuaili@cse.cuhk.edu.hk
}
\date{}
\maketitle

\begin{abstract}
We consider a new setting of online clustering of contextual cascading bandits, an online learning problem where the underlying cluster structure over users is unknown and needs to be learned from a random prefix feedback. More precisely, a learning agent recommends an ordered list of items to a user, who checks the list and stops at the first satisfactory item, if any. We propose an algorithm of \textit{CLUB-cascade} for this setting and prove a $T$-step regret bound of order $\tilde{O}(\sqrt{T})$. Previous work \cite{li2016contextual} corresponds to the degenerate case of only one cluster, and our general regret bound in this special case also significantly improves theirs. We conduct experiments on both synthetic and real data, and demonstrate the effectiveness of our algorithm and the advantage of incorporating online clustering method.
\end{abstract}

\section{1. Introduction}

Most recommendation systems nowadays display items in an ordered list. Examples include typical hotels/restaurants/goods recommendation, search engines, etc. This is especially the case for apps or games recommendations on mobile devices due to the limited size of screen. Click behaviors and feedback in such ordered lists have their distinctive features, and a {\it cascade model} was recently developed for studying feedback of user click behaviors \cite{craswell2008experimental}. In the model, after a user receives a list of items, she checks the items in the given order and clicks the first satisfactory one. After the click, she stops checking the rest items in the list. The learning agent receives the feedback of the click and knows that the items before the clicked one have been checked and are unsatisfactory, but whether the user likes any items \textit{after} the clicked one is unknown. The cascade model is straightforward but effective in characterizing user behaviors \cite{chuklin2015click}.

In this paper, we consider an online learning variant of {\it cascading bandits} \cite{kveton2015cascading,kveton2015combinatorial}. In our model, the learning agent uses exploration-exploitation techniques to learn the preferences of users over items by interactions with users. At each time step, the learning agent recommends a list of items to the current user, observes the click feedback, and receives a reward of $1$ if the user clicks on an item (and receives reward $0$ otherwise). The learning agent aims to maximize its cumulative rewards after $n$ rounds. Previous work \cite{li2016contextual,zong2016cascading} considered a setting of {\it linear cascading bandits} to deal with ever-changing set of items. Roughly speaking, the learning agent adaptively learns a linear mapping between expected rewards and features of items and users.

One important limit of the linear cascading bandit algorithms is that they mainly work in a content-dependent regime, discarding the often useful method of collaborative filtering. One way to utilize the collaborative effect of users is to consider their clustering structure. In this paper, we formulate the problem of online clustering of contextual cascading bandits, and design an algorithm to learn the clustering information and extract user feature vectors adaptively with low cumulative regret. Following the approach in \cite{gentile2014online}, we use a dynamic graph on all users to represent clustering structure, where an edge indicates the similarity between the two users. Edges between different clusters are gradually removed as the algorithms learns from the feedback that the pairs of users are not similar. We prove an upper bound of $O(d\sqrt{mKT}\ln(T))$ for the cumulative regret, where $m$ is the number of clusters, $d$ is the dimension of feature space, $n$ is the number of rounds, and $K$ is the number of recommended items. This extends and improves the existing results in the degenerate setting of only one cluster ($m=1$). Finally we experiment on both synthetic and real datasets to demonstrate the advantage of the model and algorithm.

The organization of this paper is as follows. We first introduce previous work related to our setting, then formulates the setting of {\it Online Clustering of Contextual Cascading Bandits} with some appropriate assumptions. Next we give our UCB-like algorithm CLUB-cascade and the cumulative regret bound, which is better than the existing results in the degenerate case. Then we report experimental results on both synthetic data and real data to demonstrate the advantage of incorporating online clustering. Last is the conclusion of the paper.
\section{2. Related Work}
	
The work \cite{kveton2015cascading,katariya2016dcm} introduced the click model with cascading feedback and DCM feedback to the MAB framework, which describes the random feedback dependent on the display order of items. In the cascading feedback, a user clicks the first satisfying items and stops checking further, while in the DCM feedback, after a user clicks an item, there is a chance that she is not satisfied and continues checking. \cite{kveton2015combinatorial} considered the problem where the random feedback stops at the first default position (reward $0$), in comparison with the first success position (reward $1$) in the cascade setting. Even though the settings are similar, the techniques are totally different because of the asymmetry of the binary OR function and binary AND function. \cite{zoghi2017online} brought up an online elimination algorithm to deal with different click models. All the above works focused on the setting of fixed item set.

The work \cite{li2016contextual} generalized both the cascade setting and combinatorial cascade setting with contextual information, position discounts and more general reward functions. For the binary OR case, they provided a regret bound for $T$ rounds with order $O(\frac{d}{p^\ast}\sqrt{KT}\ln(T))$ where $p^\ast$ is probability to check all recommended items and could be small. At the same time, \cite{zong2016cascading} also generalized the cascade setting with linear payoff and brought up a UCB-like algorithm, CascadeLinUCB, as well as a Thompson sampling (TS) algorithm without a proof. They proved a regret bound of $T$ rounds for the CascadeLinUCB algorithm of order $O(dK\sqrt{T}\ln(T))$. In this paper, we consider the basic cascade setting, where the random feedback stops at the first click position, together with the online clustering to explore user structure. We provide a regret bound of order $O(d\sqrt{mKT}\ln(T))$. Cast in this framework, the existing results studied the degenerate case of $m = 1$.

The work \cite{gentile2014online} first considered online clustering of linear bandits, and maintained a graph among users and used connected components to denote user clusters. A follow-up \cite{li2016collaborative} explored item structures to help cluster users and to improve recommendation performance. \cite{gentile2017context} considered a variant where the clusters over users are dependent on the current context. In this paper, we employ some idea of the first paper, and manage to make it to work with random feedback in click models. \cite{combes2015learning} considered a similar setting of clustered users with cascade feedback, but in their paper, the clusters are fixed and known to the learning agent. In our paper, the cluster structure is unknown and has to be learned by the learning agent.
\section{3. Problem Setup}

In this section, we formulate the problem of ``Online Clustering of Contextual Cascading Bandits''. In this problem, there are $u$ users, denoted by set $[u] = \{1, \ldots, u\}$. At each time step $t$, a user $i_t$ comes to be served with contents and the learning agent receives the user index with a finite feasible content set $D_t \subset \RR^{d\times 1}$, where $\norm{x}_2  \le 1$ for all $x \in D_t$. Then the learning agent recommends a ranked list of {\it distinct} $K$ items $X_t = (x_1, ..., x_K) \in \Pi^K(D_t)$ to the user. The user checks the items in the order from the first one to the last one, clicks the first attractive item, and stops checking after the click. We use the Bernoulli random variable $\by_{t,k}$ to indicate whether the item $x_{t,k}$ has been clicked or not. The learning agent receives the feedback of the index of the clicked item, that is
\begin{equation}
	\bC_{t} = \inf\{k: \by_{t,k} = 1\}.
\end{equation}
Note that $\inf(\emptyset) = \infty$ and $\bC_t = \infty$ represents that the user does not click on any given item. Let 
$
	\bK_t = \min\{\bC_t, K\}.
$
The user checks the first $\bK_t$ items and the learning agent receives the feedback $\{\by_{t,k}, k=1,\ldots,\bK_t\}$.

Let $\cH_t$ be the entire history information until the end of round $t$. Then the action $\bX_t$ is $\cH_{t-1}$-adaptive. We will write $\EE_t[\cdot]$ for $\EE[\cdot|\cH_{t-1}]$ for convenience of notation, use the boldface symbols to denote random variables, and denote $[m] = \{1,\ldots,m\}$. 

We assume the probability of clicking on an item to be a linear function of item feature vector. Specifically there exists a vector $\theta_{i_t} \in \RR^{d \times 1}, \norm{\theta_{i_t}}_2 \le 1$ for user $i_t$, such that the expectation of the binary click feedback $\by$ on the checking item $x$ is given by the inner product of $x$ with $\theta_{i_t}$, i.e.,
\begin{equation}
	\EE_t[\by | x ] = \theta_{i_t}^{\top}x,
\end{equation}
independently of any other given item.

We assume that there are $m$ clusters among the users, where $m\ll u$, and the partition of the clusters is fixed but unknown. Specifically we use $I_1, \ldots, I_m$ to denote the true clusters and a mapping function $j: [u] \to [m]$ to map user $i$ to its true cluster index $j(i)$ (user $i$ belongs to cluster $I_{j(i)}$). We assume the order of user appearance and the set of feasible items are not under the control of the learning agent. In addition, we assume clusters, users, and items satisfy the following assumptions.

\paragraph{Cluster regularity} All users in the same cluster $I_j$ share the same $\theta$, denoted as $\theta_j$. Users in different clusters have a gap between their $\theta$'s, that is
$$
	\norm{\theta_j - \theta_{j'}} \ge \gamma >0
$$
for any $j\ne j'$.

\paragraph{User uniformness} At each time step $t$, the user is drawn uniformly from the set of all users $[u]$, independently over past.

\paragraph{Item regularity} At each time step $t$, given the size of $D_t$, the items in $D_t$ are drawn independently from a fixed distribution $\bx$ with $\norm{\bx}_2 \le 1$, and $\EE[\bx\bx^{\top}]$ is full rank with minimal eigenvalue $\lambda_x > 0$. Also at all time $t$, for any fixed unit vector $\theta \in \RR^d$, given the size of $D_t$, $(\theta^{\top}X)^2$ has sub-Gaussian tail with variance parameter $\sigma^2 \le \lambda_x^2 /(8\log(4|D_t|))$.

Note that the above assumptions follow the settings from previous work \cite{gentile2014online}. 
We will have more discussions on these in later section.

At each time step $t$, the reward of $X = (x_1, \ldots, x_K) \in \Pi^K(D_t)$ under the random click result $\by_t = (\by_t(x))_{x\in D}$ (if known) is
$$
	f(X, \by_t) = 1 - \prod_{k=1}^K (1-\by_t(x_k)).
$$
By independence assumption, it is easily verified that the expectation of $f(X,\by_t)$ is
$$
	f(X, y_t) = 1 - \prod_{k=1}^K (1-y_t(x_k)),
$$
where $y_t(x) = \theta_{j(i_t)}^{\top} x$ and $y_t = (y_t(x))_{x\in D_t}$. Let
$$
	X_t^{\ast} = \argmax_{X \in \Pi^K(D_t)} f_t(X, y_t)
$$
be the optimal action in round $t$. Then the regret in time step $t$ is 
$$
	R_t(X, \by_t) = f(X_t^{\ast}, \by_t) - f(X, \by_t).
$$
The goal for the learning agent is to minimize the expected cumulative regret
\begin{equation}
	R(T) = \EE\Big[\sum_{t=1}^T R_t(\bX_t, \by_t)\Big].
\end{equation}
\section{4. Algorithm and Results}

\begin{algorithm}[tbh!] 
\caption{CLUB-cascade}
\label{alg:ClubCascade}
\begin{algorithmic}[1]
\STATE {{\bf Input:} $\lambda,\alpha, \beta>0$}
\STATE {{\bf Initialize:} $G_0 = ([u],E_0)$ is a complete graph over all users, $\bS_{i,0} = 0_{d\times d}, \bb_{i,0} = 0_{d \times 1}, T_{i,0}=0$ for all $i \in [u]$. \label{alg:clubcascade:initialize} }

\FORALL {$t=1,2,\ldots, T$}
\STATE {Receive user index $i_t$, and the feasible context set $D_t \subset \RR^{d\times 1};$ \label{alg:clubcascade:receive user index} }
\STATE {\label{alg:clubcascade:find V} Find the connected component $V_t$ for user $i_t$ in the current graph $G_{t-1} = ([u], E_{t-1})$, and compute 
\begin{align*}
\bM_{V_t,t-1} &= \lambda I + \sum_{i \in V_t} \bS_{i,t-1}, \quad \bb_{V_t,t-1} = \sum_{i \in V_t} \bb_{i,t-1}, \\
\hat{\btheta}_{V_t,t-1} &= \bM_t^{-1}\bb_t; 
\end{align*}
}
\STATE {\label{alg:clubcascade:compute ucbs} For all $x \in D_t$, compute
\begin{equation}
\label{eq:Ut}
\bU_t(x) = \min\{ \hat{\btheta}_{V_t,t-1}^{\top}x + \beta \sqrt{x^{\top} \bM_{V_t,t-1}^{-1} x}, 1 \};
\end{equation}}
\STATE {\label{alg:clubcascade:recommend and receive feedback} Recommend a list of $K$ items $\bX_t = (\bx_{t,1},\ldots,\bx_{t,K})$ with largest $\bU_t(\cdot)$ values and receive feedback $\bC_t \in \{1,\ldots,K,\infty\}$;}
\STATE {\label{alg:clubcascade:update1} Update statistics
\begin{align*}
\bS_{i_t,t} &= \bS_{i_t,t-1} + \sum_{k=1}^{\bK_t} \bx_{t,k}\bx_{t,k}^{\top}, \\
\bb_{i_t,t} &= \bb_{i_t,t-1} + \sum_{k=1}^{\bK_t} \bx_{t,k} \bOne\{\bC_t=k\},\\
T_{i_t,t} &= T_{i_t,t-1} + \bK_t,
\end{align*}
where $\bK_t = \min\{\bC_t,K\}$ and update
$$
\hat{\btheta}_{i_t,t} = (\lambda I + \bS_{i_t,t})^{-1} \bb_{i_t,t};
$$}
\STATE {Update \label{alg:clubcascade:update2}
\begin{align*}
\bS_{\ell,t} &= \bS_{\ell,t-1}, \bb_{\ell,t} = \bb_{\ell,t-1}, T_{\ell,t} = T_{\ell,t-1}, \\
\hat{\btheta}_{\ell,t} &= \hat{\btheta}_{\ell,t-1}
\end{align*}
for all $\ell \ne i_t$;}
\STATE {\label{alg:clubcascade:delete} Delete the edge $(i_t, \ell) \in E_{t-1}$, if
\begin{align*}
&\norm{\hat{\btheta}_{i_t,t} - \hat{\btheta}_{\ell,t}}_2 \\
&\quad \ge \alpha \left(\sqrt{\frac{1 + \ln(1+T_{i_t,t})}{1 + T_{i_t,t}}} + \sqrt{\frac{1 + \ln(1+T_{\ell,t})}{1 + T_{\ell,t}}} \right)
\end{align*}
and obtain a new graph $G_t = ([u], E_t)$;}
\ENDFOR {~~$t$}
\end{algorithmic}
\end{algorithm}

\subsection{Notations}

Our main algorithm is given in Algorithm \ref{alg:ClubCascade}. Before diving into details, let us define some useful notations used in later analysis. For any time step $t$ and user $i$, define
\begin{align*}
\bS_{i,t} &= \sum_{ \substack{ s\leq t \\ i_s=i } } \sum_{k=1}^{\bK_s} \bx_{s,k}\bx_{s,k}^{\top},\quad \bb_{i,t} = \sum_{ \substack{ s\leq t \\ i_s=i } } \sum_{k=1}^{\bK_s} \by_{s,k}\bx_{s,k}, \\
T_{i,t} &= \sum_{ s\leq t,\ i_s=i  } \bK_s
\end{align*}
to be the Gramian matrix, the moment matrix of regressand by regressors, and the number of effective feedbacks for user $i$ up to time $t$, respectively. Let $\emptyset \neq I \subset [u]$ be any nonempty user index subset and
\begin{equation}
\begin{split}
&\bM_{I,t} = \lambda I_d + \sum_{i \in I} \bS_{i,t}, \quad \bb_{I,t} = \sum_{i \in I} \bb_{i,t},\\
&\hat{\btheta}_{I,t} = \bM_{I,t}^{-1}\bb_{I,t}, \quad T_{I,t} = \sum_{i\in I} T_{i,t}
\end{split}
\end{equation}
be the regularized Gramian matrix, the moment matrix of regressand, the estimate by ridge regressors, and the frequency associated with user set $I$ and regularization parameter $\lambda >0$ up to time $t$, respectively.

\subsection{Algorithm}

The algorithm maintains an undirected graph structure on all users $G_t = ([u], E_t)$, where an edge exists between a pair of users if they are similar. The collection of the connected components represents a partition of the users. 

The learning agent starts with a complete graph over all users and initializes Gramian matrix and the moment matrix of regressand for each user $i$ (Line \ref{alg:clubcascade:initialize}). At each time step $t$, the learning agent receives a user index $i_t$ and a feasible finite content set $D_t$ (Line \ref{alg:clubcascade:receive user index}), where $\norm{x}_2 \le 1$ for all $x \in D_t$. From the current graph structure on users $G_{t-1} = ([u], E_{t-1})$, the agent finds the connected component $V_t$ containing user $i_t$ and computes the Gramian matrix, the moment matrix of regressand, and the estimates $\hat{\btheta}_{V_t,t-1}$ by ridge regressor associated with set $V_t$ up to time $t-1$ (Line \ref{alg:clubcascade:find V}). Then it uses this $\hat{\btheta}_{V_t,t-1}$ as the estimate for the true weight vector $\theta_{j(i_t)}$ to compute the upper confidence bound of the expected reward $\theta_{j(i_t)}^{\top}x$ for each item $x \in D_t$ (Line \ref{alg:clubcascade:compute ucbs}). This step relies on the following lemma, which gives the theoretical guarantee of the ridge regression estimate for the true weight vector.

\begin{lemma}
\label{lem:thetaEstimate}
Suppose $(x_1, y_1), \ldots, (x_t, y_t), \ldots$ are generated sequentially from a linear model such that $\norm{x_t} \le 1$ for all $t$, $\EE[y_t | x_t] = \theta_{\ast}^{\top} x_t$ for fixed but unknown $\theta_{\ast}$ with norm at most $1$, and $\{y_t - \theta_{\ast}^{\top} x_t\}_{t=1,2,\ldots}$ have $R$-sub-Gaussian tails. Let $M_t = \lambda I + \sum_{s=1}^{t} x_s x_s^\top, b_t = \sum_{s=1}^{t} x_s y_s$, and $\delta > 0$. If $\hat{\theta}_t = M_t^{-1} b_t$ is the ridge regression estimator of $\theta_{\ast}$, then with probability at least $1-\delta$, for all $t \ge 0$,
\begin{align*}
\norm{\hat{\theta}_t - \theta_{\ast}}_{M_t} &\le R \sqrt{d\ln\left(1 + \frac{t}{\lambda d}\right) + 2\ln\frac{1}{\delta} } + \sqrt{\lambda} \\
&=: \beta(t,\delta)\,.
\end{align*}
\end{lemma}
This Lemma is by \cite[Theorem 2]{abbasi2011improved}.

When the current cluster is correct (which is guaranteed after $O(\ln(T))$ rounds and to be proved later), i.e. $V_t = I_{j(i_t)}$, 
$$
\norm{\theta_{j(i_t)} - \hat{\btheta}_{V_t,t-1}}_{\bM_{V_t,t-1}} \le \bbeta(T_{V_t,t-1},\delta) \le \bbeta(n,\delta).
$$
(Here for a positive-definite matrix $M$, define the norm $\|x\|_M = \sqrt{x^T M x}$. It is not hard to verify that if $M \succ 0$, the dual norm $\|x\|_M$ is $\|x\|_{M^{-1}}$.) Then by Cauchy-Schwarz inequality, we have
\begin{align*}
&\left| \hat{\btheta}_{V_t,t-1}^{\top}x - \theta_{j(i_t)}^{\top}x \right| \\
\le& \norm{\hat{\btheta}_{V_t,t-1} - \theta_{j(i_t)}}_{\bM_{V_t,t-1}} \norm{x}_{\bM_{V_t,t-1}^{-1}} \\
\le& \bbeta(T,\delta) \norm{x}_{\bM_{V_t,t-1}^{-1}}\,,
\end{align*}
which results in a confidence interval for the expected reward $\theta_{j(i_t)}^{\top}x$ on each item $x\in D_t$.

Next the learning agent recommends a list of $K$ items $\bX_t = (\bx_1,\ldots,\bx_K)$ which have the largest upper confidence bounds. The user $i_t$ checks the recommended items from the first one, clicks on the first satisfactory item, and stops checking anymore. Then the learning agent receives feedback $\bC_t \in \{1,\ldots,K,\infty\}$ (Line \ref{alg:clubcascade:recommend and receive feedback}). $\bC_t \in\{1,\ldots,K\}$ means that the user clicks $\bC_t$-th item and the first $\bC_t-1$ items are not satisfactory, while the items after $\bC_t$-th position are not checked by the user. $\bC_t = \infty$ means that the user has checked all recommended items but none of them is satisfactory. Based on the feedback, the learning agent updates its statistics on user $i_t$ (Line \ref{alg:clubcascade:update1}) but not on other users (Line \ref{alg:clubcascade:update2}).

Based on the updates, the weight vector estimate for the user $i_t$ might change and the similarity with other users might be verified false. The learning agent checks the edge of $(i_t,\ell) \in E_{t-1}$ for any user $\ell$ that is linked to user $i_t$ and deletes it if the distance between the two estimated weight vectors is large enough (Line \ref{alg:clubcascade:delete}).

\subsection{Analysis}

The following theorem gives a bound on the cumulative regret achieved by our algorithm CLUB-cascade.

\begin{theorem}
\label{thm:main}
Suppose the cluster structure on the users, user appearance, and items satisfy the assumptions stated in the section of Problem Setup with gap parameter $\gamma>0$ and item regularity parameter $0<\lambda_x\le 1$. Let $\lambda,K$ be the regularization constant and the number of recommended items in each round. Let $\lambda \ge K, \beta = \sqrt{d\ln\left( 1+\frac{T}{\lambda d} \right) + 2\ln(4mT)}+\sqrt{\lambda}$ and $\alpha = \sqrt{32d/\lambda_x}$, where $d,m,u$ denotes the feature dimension, the number of clusters and the number of users, respectively. Then the cumulative regret of CLUB-cascade algorithm for $T$ rounds satisfies
\begin{align}
R(T) \le & 2 \left(\sqrt{d\ln\left(1+\frac{T}{\lambda d}\right) + 2\ln(4mT)} + \sqrt{\lambda}\right) \notag\\
&\quad \cdot \sqrt{2dmKT\ln\left(1+\frac{TK}{\lambda d}\right)} \notag\\
&\qquad + O\left( u\left(\frac{d}{\gamma^2 \lambda_x} + \frac{1}{\lambda_x^2}\right) \ln(T) \right)\\
\le &O\left( d\sqrt{mKT}\ln(T) \right). \notag
\end{align}
\end{theorem}

For the degenerate case when $m=1$, our result improves the existing regret bounds.

\begin{corollary}
\label{cor:m=1}
When the number of clusters $m=1$, that is all users are treated as one, let $\lambda = K$ and $\beta = \sqrt{d\ln\left( 1+\frac{T}{\lambda d} \right) + 2\ln(4T)}+\sqrt{\lambda}$. Then the cumulative regret of CLUB-cascade after $n$ rounds satisfies
\begin{align}
R(T) \le &2 \left(\sqrt{dT\ln\left(1+\frac{T}{\lambda d}\right) + 2\ln(4T)} + \sqrt{\lambda}\right)\notag \\
&\quad \cdot \sqrt{2dK\ln\left(1+\frac{TK}{\lambda d}\right)} \notag\\
\le &O(d\sqrt{KT}\ln(T)).
\end{align}
\end{corollary}

Note this result improves the existing results\cite{li2016contextual,zong2016cascading}. Discussions about the results, problem assumptions and implementations are given later.

Next we give a proof sketch for the Theorem \ref{thm:main}.

\begin{proof}[Sketch for Theorem \ref{thm:main}]
The proof for the main theorem is mainly based on two parts. The first part proves the exploration rounds needed to guarantee the clusters partitioned correctly. And the second part is to estimate regret bounds for linear cascading bandits after the clusters are partitioned correctly.

Under the assumption of item regularity, we prove when $T_{i,t} \ge O\left(\frac{d}{\gamma^2 \lambda_x}\ln(T)\right)$, the $\norm{\cdot}_2$ confidence radius for weight vector associated with user $i$ will be smaller than $\gamma/2$, where the $\gamma$ is the gap constant raised in the assumption of cluster regularity. Suppose user $i$ and user $\ell$ belong to different clusters and the effective number of feedbacks associated to both user $i$ and $\ell$ meet the requirements. Then the condition in the Algorithm \ref{alg:ClubCascade} of deleting an edge $(i, \ell)$ (Line 10) will be satisfied, thus the edge between user $i$ and $\ell$ will be deleted under our algorithm with high probability. On the other hand, if the condition of deleting an edge $(i, \ell)$ is satisfied, then the $\norm{\cdot}_2$ difference between the weight vectors is greater than $0$, thus the two users belong to different clusters, by the assumption of cluster regularity.

By the assumption of item regularity and Bernstein's inequality, after
\begin{align*}
t \ge O\left( u\left(\frac{d}{\gamma^2 \lambda_x} + \frac{1}{\lambda_x^2}\right) \ln(T) \right)
\end{align*}
rounds, we could gather enough information for every user, thus resulting a correct clustering with high probability.

After the clusters are correctly partitioned, the recommendation is based on the estimates of cluster weight vector with the cascade feedback collected so far. After decomposing, the instantaneous regret can be bounded by the individual difference between expected rewards of best items and checked items, which can be bounded with $2\beta \norm{x}_{\bM}$, by the definition of $\bU_t(x)$. Then it remains to bound the sum of self-normalized sequence $\sum_{t=1}^T \norm{x_t}_{M_{t-1}^{-1}}$, where $M_t = M_{t-1} + xx^{\top}$.
\end{proof}

\subsection{Extensions to Generalized Linear Rewards}

In this section, we consider a general case that the expected reward of recommending item $x$ to user $i_t$ at round $t$ is 
$$
\EE_t[\by | x ] = \mu(\theta_{i_t}^{\top}x)\,,
$$
where $\mu$ is a strictly increasing link function, continuously differentiable, and Lipschitz with constant $\kappa_{\mu}$. This definition arises from exponential family distributions \cite{filippi2010parametric} and incorporates a large class of problems, like Poisson or logistic regression. Let $c_{\mu} = \inf_{a \in [-2,2]} \mu'(a)$ and assume $c_{\mu} > 0$.

In this setting, let the estimator $\hat{\btheta}_{I,t-1}$ for the set of users $I$ be maximum likelihood estimator, or equivalently \cite{filippi2010parametric,li2017provably} the unique solution of 
\begin{equation}
\sum_{s=1}^{t-1} \bOne\{i_s \in I\} \sum_{k=1}^{\bK_s} \left( \by_{s,k} - \mu(\theta^\top \bx_{s,k}) \right) \bx_{s,k} = 0,
\end{equation}
which can be found efficiently using Newton's algorithm. Note that the original samples $(\bx_{s,k}, \by_{s,k})$ are stored instead of only aggregation $\bS,\bb$ in the linear case. With a slightly modified version of Algorithm \ref{alg:ClubCascade}, a result of the cumulative regret bound is obtained and provided in the following theorem.

\begin{theorem}
\label{thm:glm}
Under the same assumptions and notations in linear setting, let
\begin{align*}
\beta = \frac{1}{c_{\mu}}\sqrt{ \frac{8}{\lambda_x} + d\ln(T/d)+ 2\ln(4mT)}
\end{align*}
and $\alpha = \sqrt{32d/(\lambda_x c_{\mu}^2)}$, where $d,m,u$ denotes the feature dimension, the number of clusters, and the number of users, respectively. Then the cumulative regret of CLUB-cascade with generalized linear rewards, after $T$ rounds, satisfies
\begin{align}
R(T) \le &2 \kappa_{\mu} \beta \sqrt{2dmKT\ln\left(1+\frac{KT}{\lambda d}\right)} \notag\\
&\quad + O\left( u\left(\frac{d}{\gamma^2 \lambda_x} + \frac{1}{\lambda_x^2}\right) \ln(T) \right) \\
\le &O\left( \frac{\kappa_{\mu}d}{c_{\mu}}\sqrt{mKT}\ln(T) \right)\,. \notag
\end{align}
\end{theorem}

\subsection{Discussions} \label{sec:discuss}

The degenerate case where the number of clusters $m=1$, or equivalently all users are treated as the same type, has the same setting with Section 4.2.2 of \cite{li2016contextual} and the setting in \cite{zong2016cascading}. The regret proved in the first paper is $O(\frac{d}{p^\ast}\sqrt{KT}\ln(T))$, which has an additional term $1/p^{\ast}$ compared to ours. The parameter $p^{\ast}$ denotes the minimal probability that a user has checked all items, which could be quite small. The reason that we can get rid of such a $1/p^{\ast}$ term is because we have a better regret decomposition formula than theirs. The regret presented in the second paper has the bound of $O(dK\sqrt{T}\ln(T))$, which has an additional term $\sqrt{K}$ than ours. This reason is that we have a tighter bound for the sum of self-normalized sequence.

For the assumption on the true cluster structure over users, we assume there is a gap $\gamma>0$ between the weight vectors associated with different clusters. The parameter $\gamma$ is a trade-off between personalization and collaborative filtering, where $\gamma = 2$ corresponds to the case of only one cluster containing all users and $\gamma$ taking the value of minimal distance between different user weight vectors corresponds to the case that each user is one cluster. Also, the assumption of $\gamma$ can be further relaxed by modifying $\gamma$ along running the algorithm. Our algorithm explores clustering structures adaptively: It starts with one cluster, then finds finer and finer clustering until the cluster distance reaches $\gamma$. As more data flows in, the parameter $\gamma$ can be changed smaller and our algorithm can continue working without the need to restart. By a similar analysis, we could derive an asymptotic regret bound (with parameters in the algorithm changed accordingly). We omit this part and simply assume a $\gamma>0$ gap exists.

For the users, we assume the learning agent has no control over user appearances and at each time step, a user is drawn uniformly from all users, and independently from the past. If the learning agent has the access to sample users, the setting becomes active learning in online clustering and should have a better regret bound because the learning agent does not need to wait for collecting enough information. The uniform user appearance assumption means that the users we take care of are on the same activity level, which is easier for us to deal with. If there is some activity structure over users, we might need further assumptions and corresponding strategies on the activity structure. For example, if there are a large amount of new users, or users who only come a few time, additional assumptions like that those users share the same prediction vectors $\theta$ might be brought up. We leave the relaxations of the two assumptions as future work.

For the items, we assume they are drawn independently from a norm $1$ vector distribution $\bx$ with $\EE[\bx\bx^{\top}]$ has minimal eigenvalue $\lambda_x>0$. This assumption is to guarantee the shrinkage of confidence ellipsoid when estimating individual weight vector, thus to distinguish different weight vectors after collecting enough information. If this assumption is violated, we can specify a barycentric spanner in the item set, as in \cite{dani2008stochastic}. \cite{lattimore2017end} also discusses the necessity of incorporating a barycentric spanner if we want to achieve efficiency in learning.

If the cluster structure on users is known, then the setting is equivalent to $m$ independent linear cascading bandits and the upper bound for the cumulative regret is $\tilde{O}(d\sum_{j=1}^m\sqrt{T_{I_j}})$, where $m$ is the number of independent linear bandits and $T_{I_j}$ is the time horizon for $j$-th linear bandits with $T_{I_1} + \cdots +T_{I_m} = T$. The upper bound reaches its maximum bound $\tilde{O}(d\sqrt{mT})$ when $T_{I_1}=\cdots=T_{I_m}=T/m$ and reaches its minimum bound $\tilde{O}(d\sqrt{T})$ when $T_{I_1} = T, T_{I_2} = \cdots = T_{I_m} = 0$.

In our algorithm CLUB-cascade, it uses connected component to represent a cluster. Thus a bigger cluster can be split into two smaller clusters only when all edges between these two smaller clusters have been deleted, which might take quite a long time. To accelerate our algorithm, a random graph initialization for all users might be adopted. To be specific, suppose all the true clusters $I_j$ satisfy $\abs{I_j} \ge \beta u/m$ and then use an Erdos-Renyi random graph as initialization, where each edge is chosen with suitable probability $p = O(m\log(u/\delta)/\beta)$. By \cite[Lemma 1]{gentile2014online}, the subgraphs of true clusters are connected with high probability.
\section{5. Experiments}

In this section, we compare our algorithm with C$^3$-UCB \cite{li2016contextual} and CascadeLinUCB \cite{zong2016cascading}, which are the most related works. In both synthetic and real datasets, the results demonstrate the advantage of incorporating online clustering in the setting of online recommendations with cascade model. We focus on linear rewards for all experiments. To accelerate our algorithm, we use a sparse initialization instead of the complete graph initialization, similar in \cite{gentile2014online}.

\subsection{Synthetic Data}

\begin{figure}
\centering
\subfigure[$\theta$ orthogonal, $m=2$]{
\includegraphics[width = 0.225\textwidth]{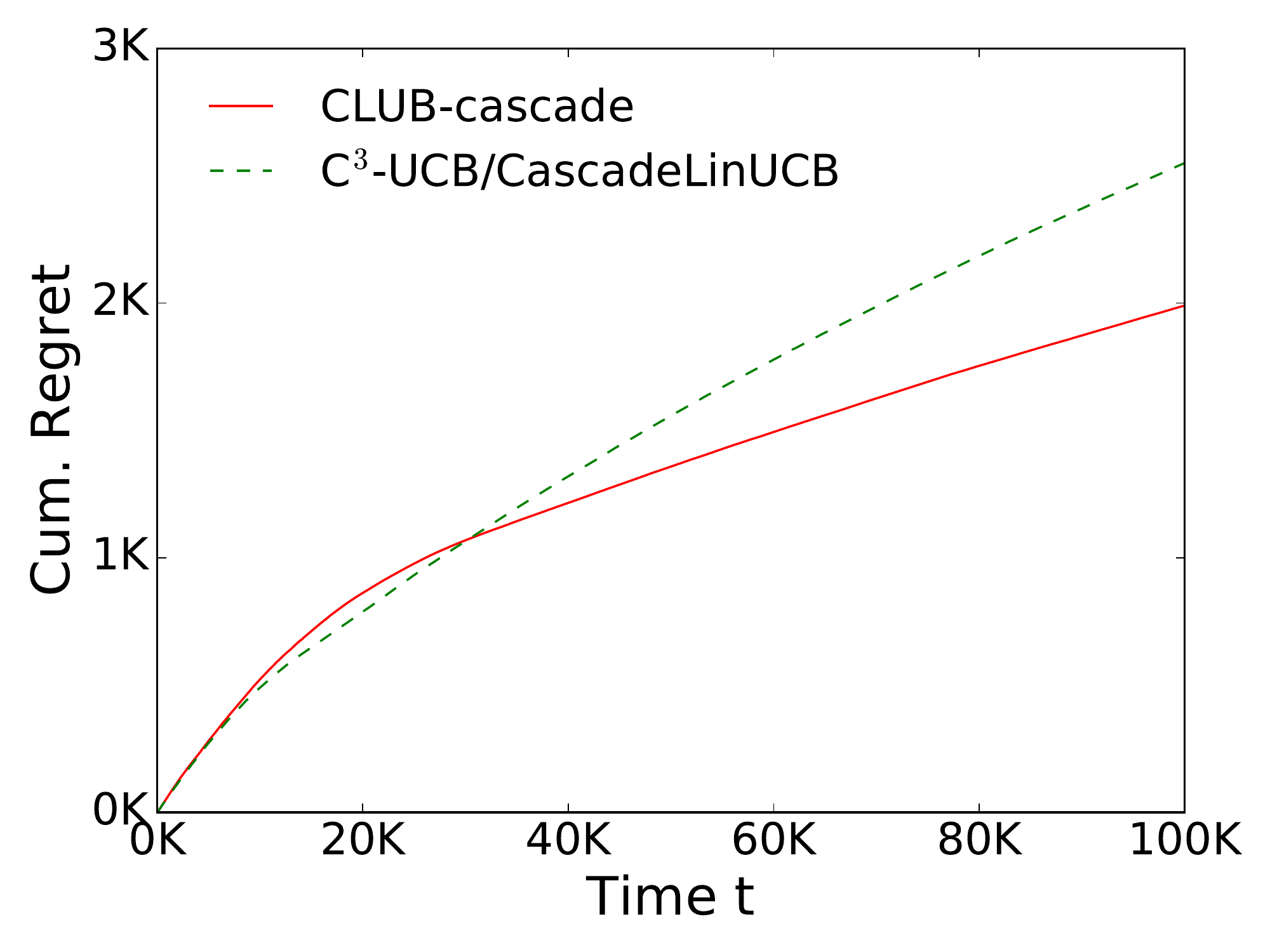}}
\subfigure[$\theta$ orthogonal, $m=5$]{
\includegraphics[width = 0.225\textwidth]{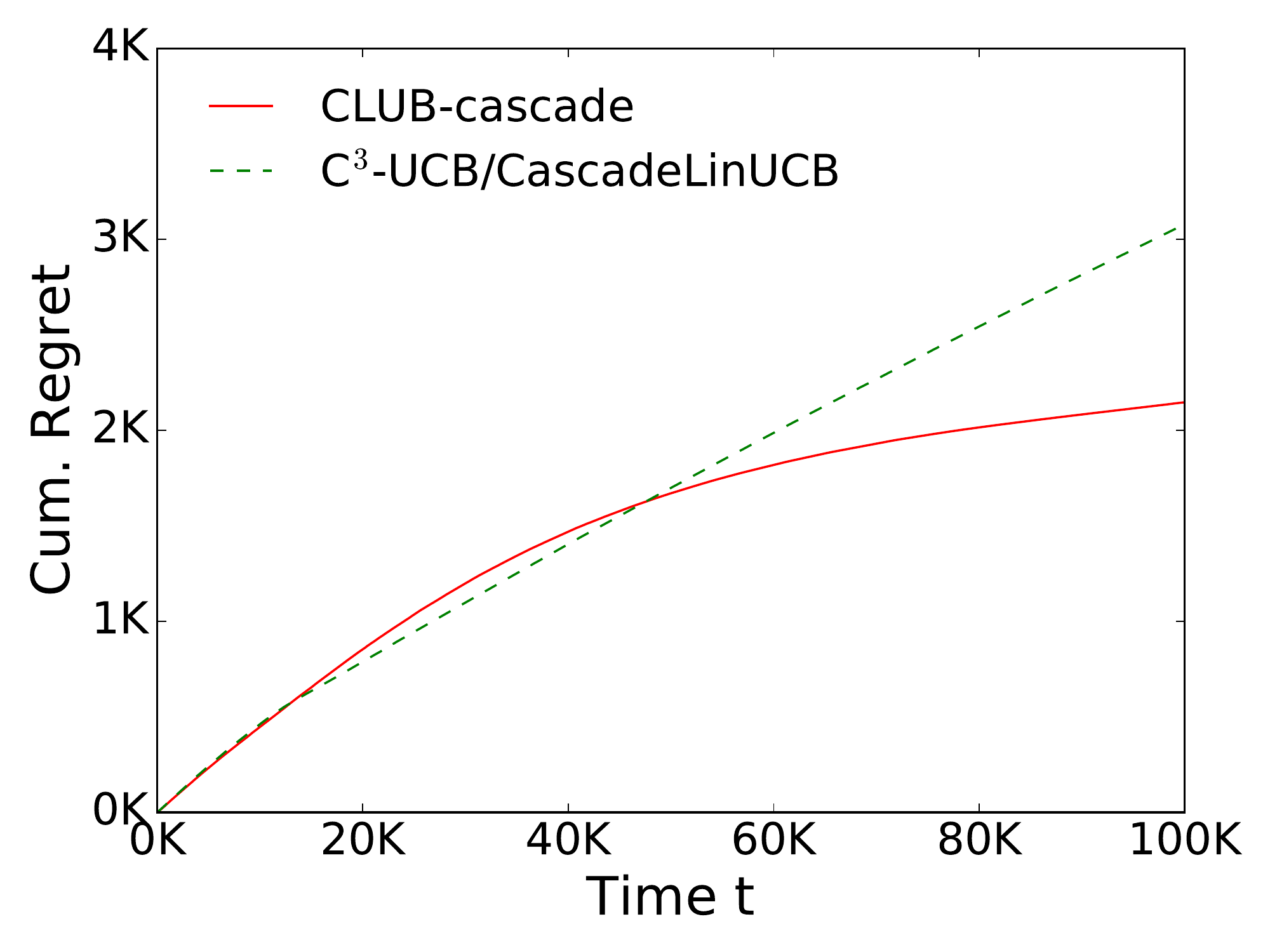}}\\
\subfigure[$\gamma = 0.2,~m=2$]{
\includegraphics[width = 0.225\textwidth]{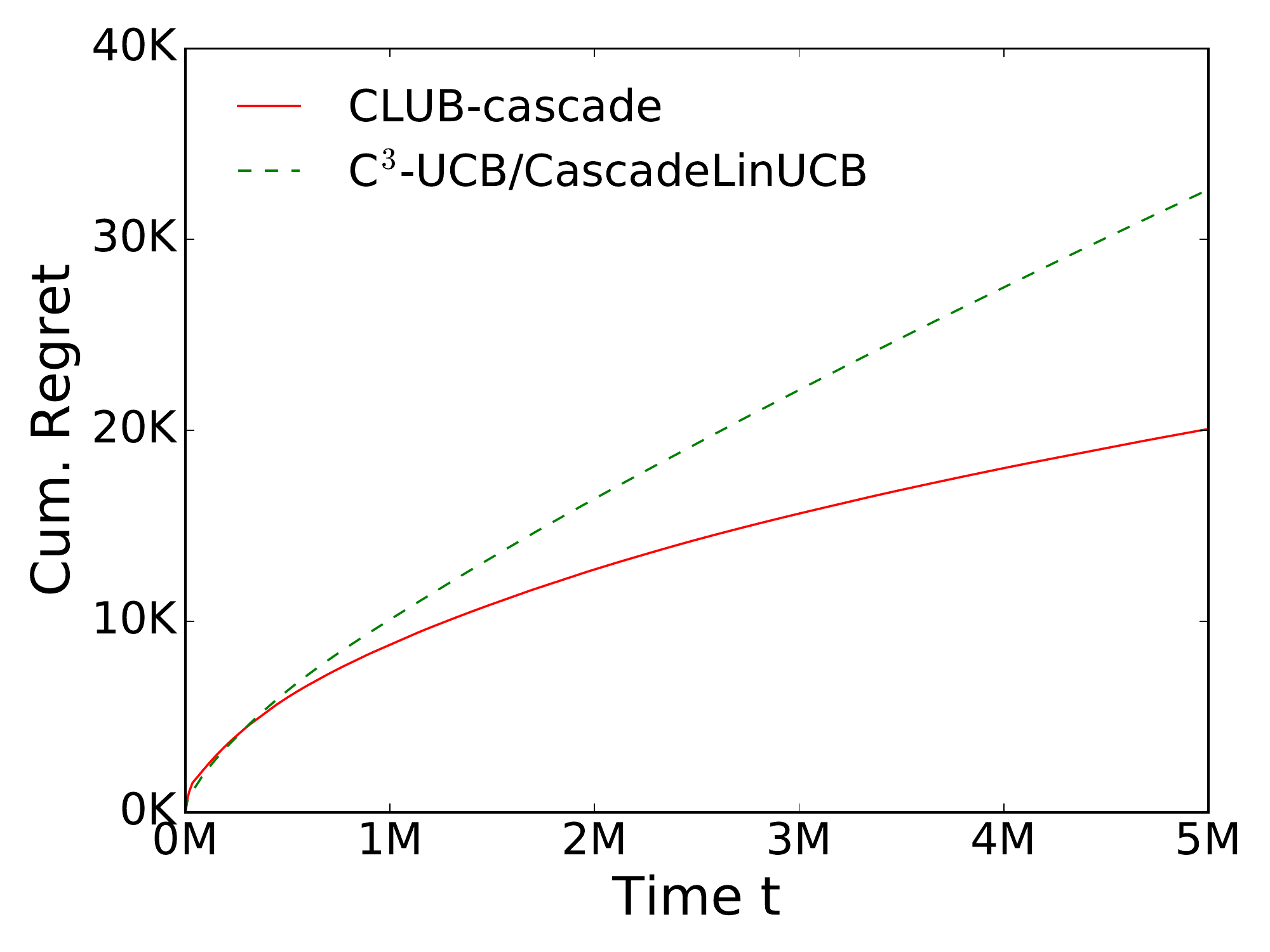}}
\subfigure[$\gamma = 0.2, m=5$]{
\includegraphics[width = 0.225\textwidth]{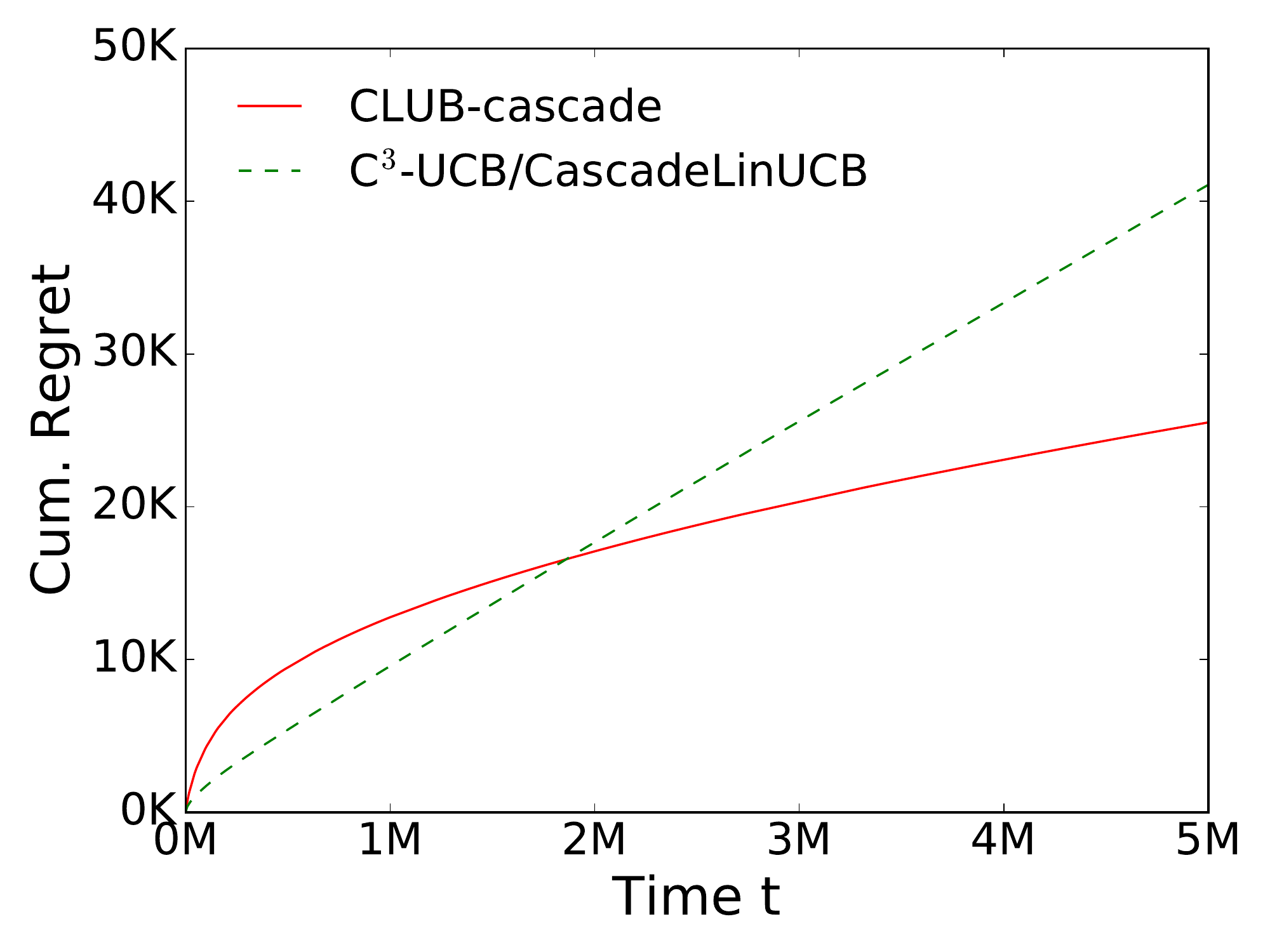}}
\caption{Synthetic Data Set, $u=40,~L=200,~K=4,~d=20$}
\label{fig:synthetic}
\end{figure}

In this section, we compare our algorithm, CLUB-cascade, with C$^3$-UCB / CascadeLinUCB on the synthetic data. The results are shown in Figure \ref{fig:synthetic}.

In all the four settings, we randomly choose a content set with $L=200$ items, each of which has a feature vector $x\in \RR^d$ with $\norm{x}_2 \le 1$ and $d=20$. We use $u=40$ users and assign them randomly to $m=2,5$ clusters. For each cluster $j\in[m]$, we fix a weight vector $\theta_j$ with norm $1$ and use it to generate Bernoulli random variable, whose mean is the inner product of $\theta_j$ with the corresponding item vector. In each round, a random user comes and the algorithm recommends $K=4$ items to the user. According to the Bernoulli random variables, the algorithm receives the cascading feedback and updates its statistics accordingly. In the synthetic setting, since we know the true weight vector $\theta_j$, the best action can be computed and thus the cumulative regret for algorithms. The vertical axis denotes the cumulative regret and the horizontal axis denotes time step $t$.

In the four subfigures, we explore the distance gap $\gamma$ between different $\theta$'s and the number of clusters $m$. When the gap $\gamma$ between weight vectors $\theta$ is fixed, our algorithm has a better advantage over theirs when the number of clusters $m$ is bigger. The $\theta$'s in subfigures (a)(c) are orthogonal, that is, the difference gap between them is $\gamma = \sqrt{2}$. The difference gap $\gamma$ in (b)(d) is set to be $0.2$, thus the cosine similarity between the different $\theta$'s is $0.98$, which is quite high. Thus under the same number of clusters, our algorithm needs more time to learn well in the setting of a smaller $\gamma$. Because $\gamma=0.2$ means near-$1$ cosine similarity, to regard all users as a whole might have advantages in early rounds. However, after our algorithm learns out the true cluster structure, their advantage depreciates very fast.

Although our algorithm needs more steps to achieve an obvious advantage with a smaller gap $\gamma$, typically it is not required to differentiate $\theta$'s with near-$1$ cosine similarity. The purpose we use this setting is to demonstrate the extreme case. In real applications, the estimated weight vectors will not be too similar so that our algorithm can easily outperform theirs, which we will see in the next experiments.

Note that our algorithm is not as good as theirs in the beginning. The reason is that we use a random sparse graph initialization and this initialization might result in inaccurate clustering for early rounds. However, after collecting enough information, our algorithm can still learn out the correct clustering (which might be a little finer for true clustering).

\subsection{Yelp Dataset}

\label{sec:yelp}

\begin{figure}
\centering
\subfigure[$u=40,~L=200$]{
\includegraphics[width = 0.225\textwidth]{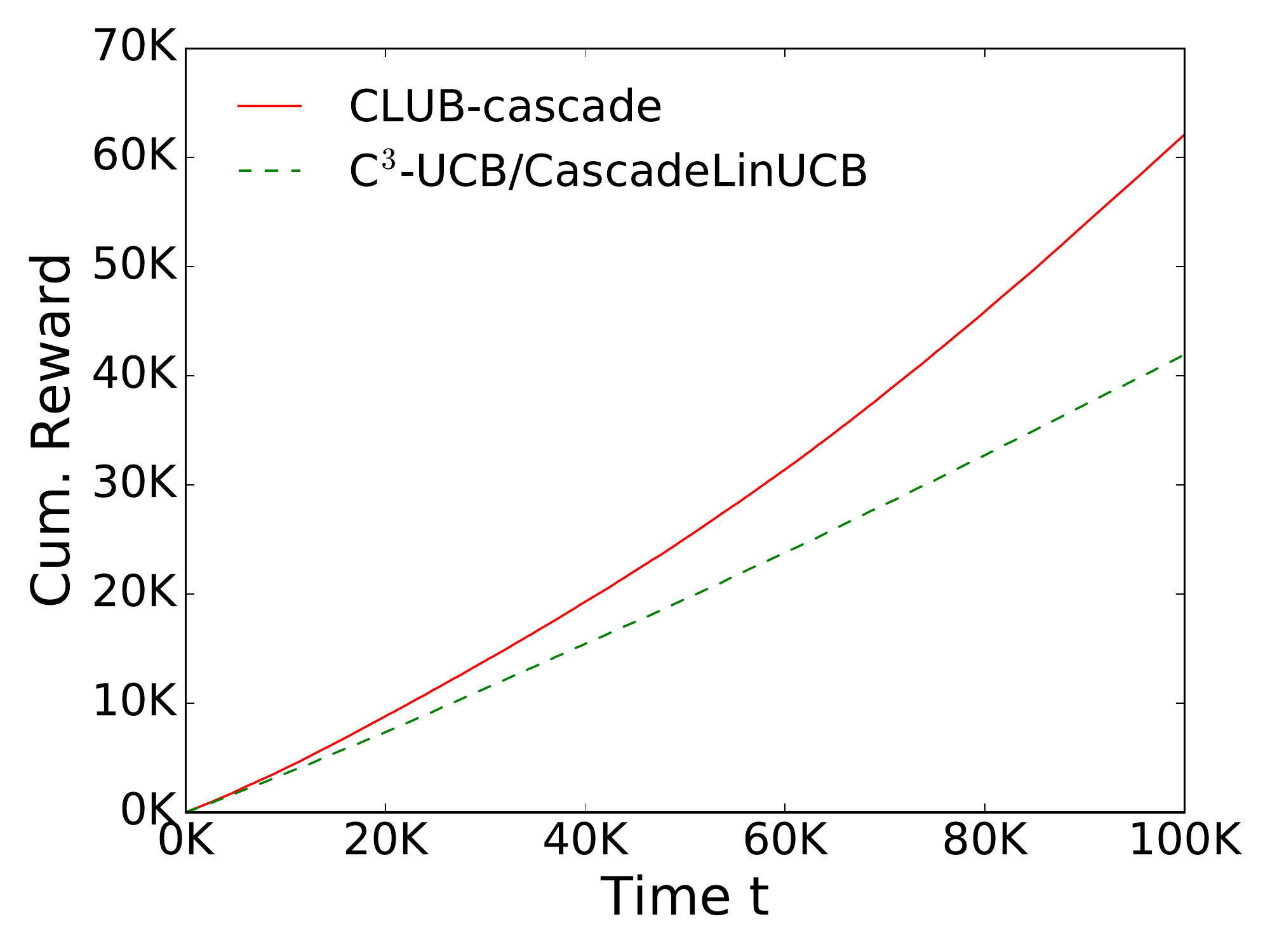}}
\subfigure[$u=40,~L=1k$]{
\includegraphics[width = 0.225\textwidth]{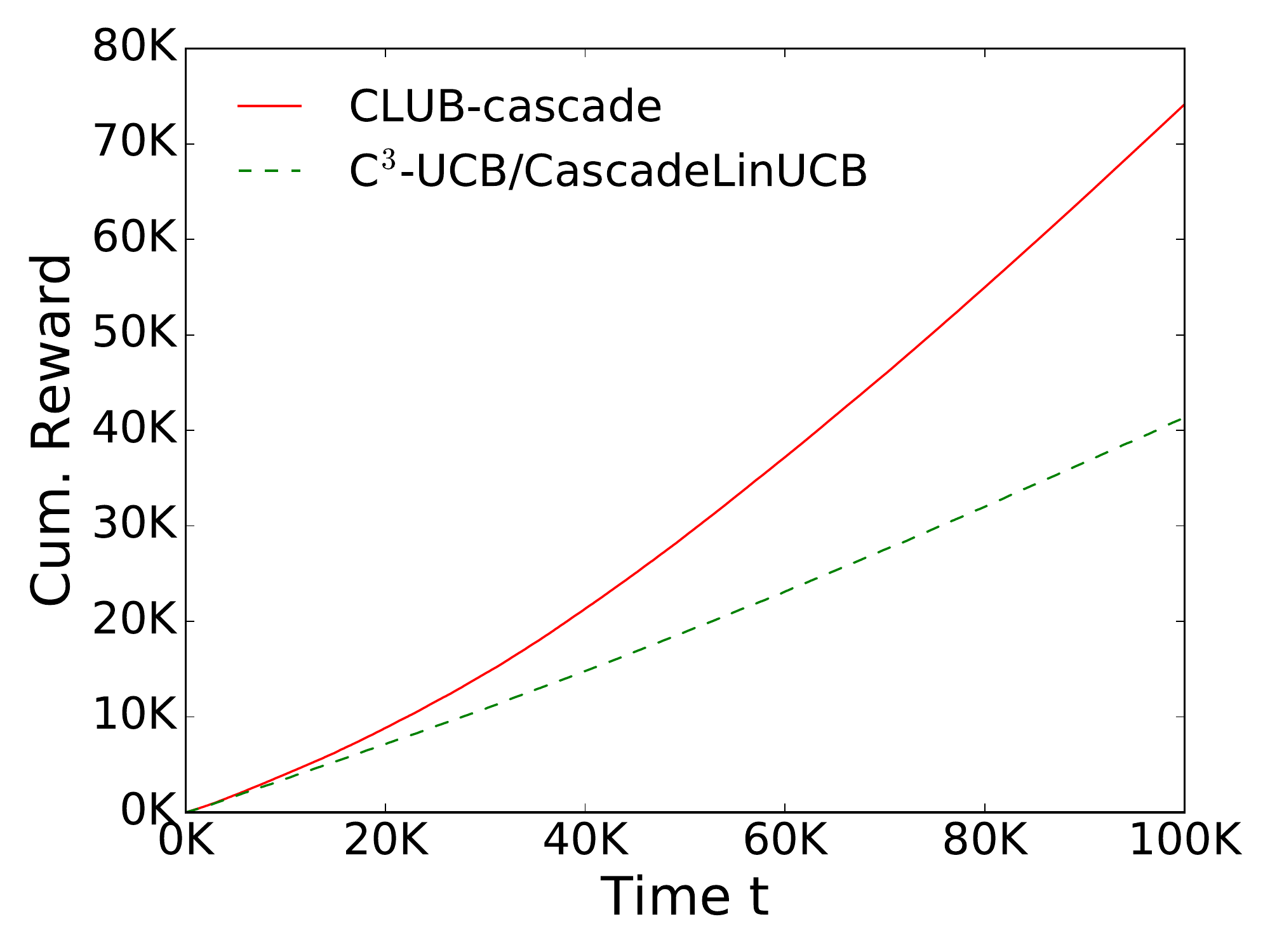}}
\subfigure[$u=200,~L=200$]{
\includegraphics[width = 0.225\textwidth]{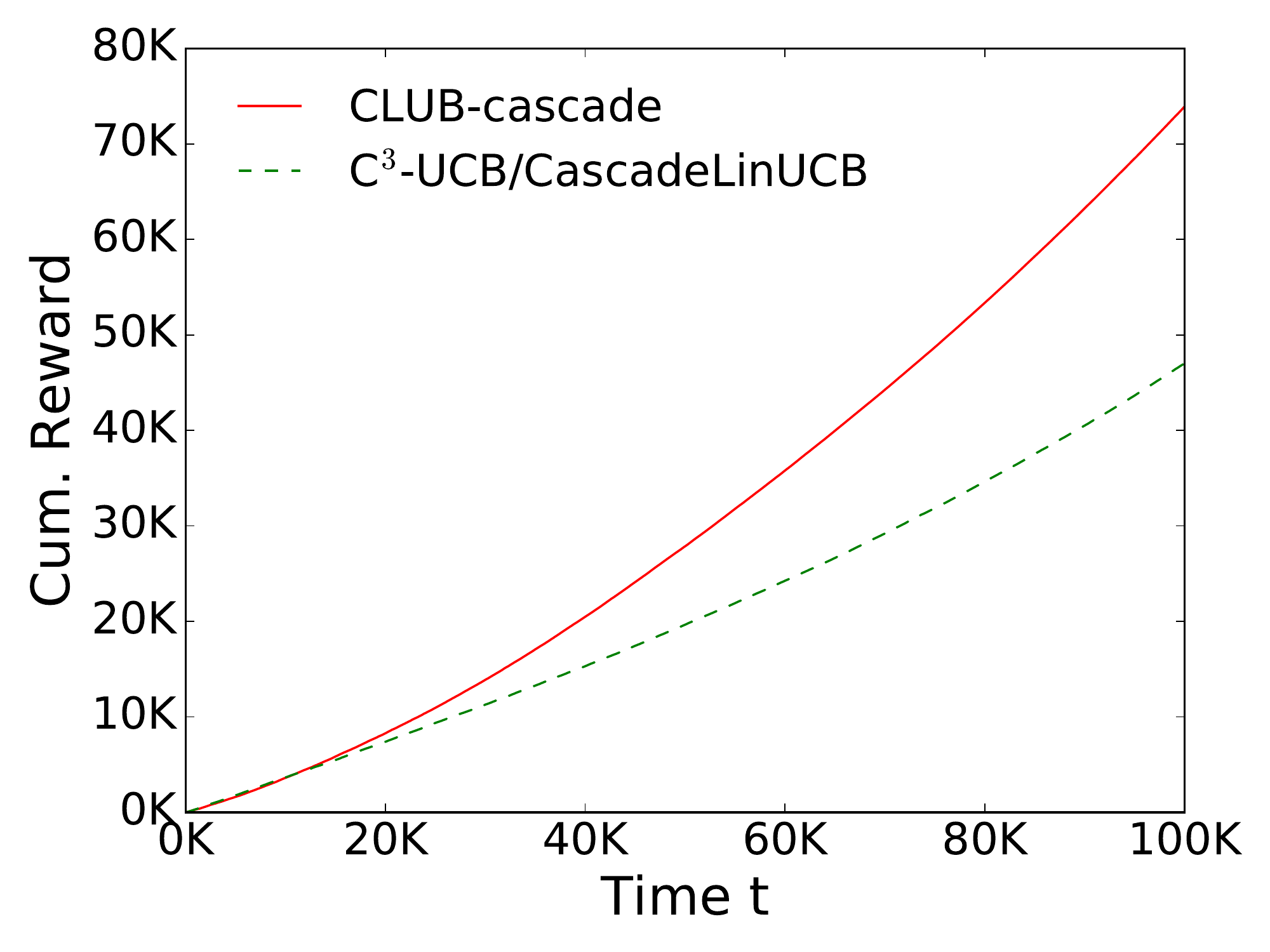}}
\subfigure[$u=200,~L=1k$]{
\includegraphics[width = 0.225\textwidth]{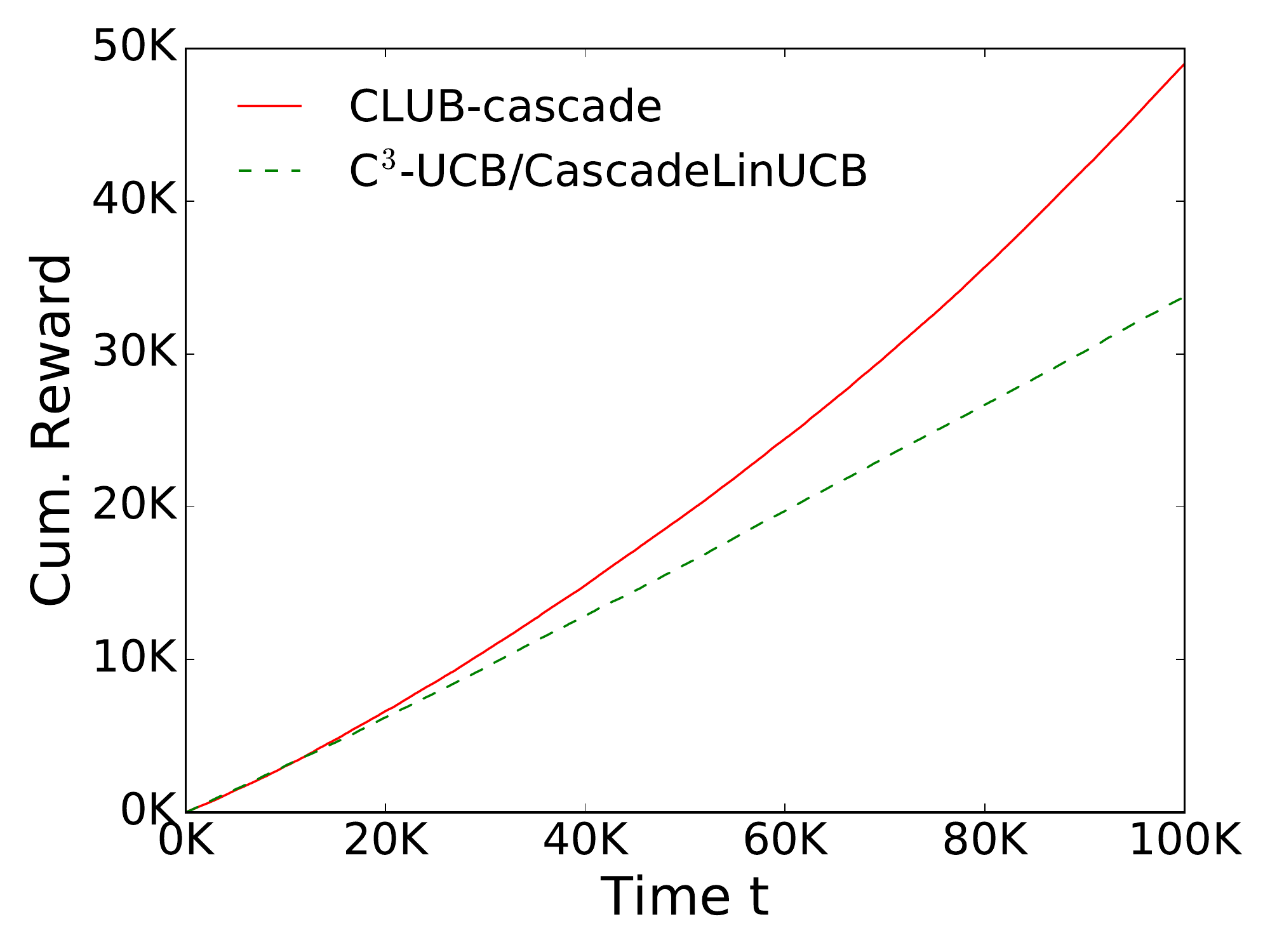}}
\caption{Cumulative clicks on Yelp dataset, $d=20, K=4$}
\label{fig:yelp_reward}
\end{figure}

In this section, we compare our algorithm, CLUB-cascade, with C$^3$-UCB / CascadeLinUCB on restaurant recommendations with Yelp dataset\footnote{$\mathrm{http://www.yelp.com/dataset\_challenge}$}. The dataset contains user ratings for several businesses. For restaurants, it contains $1,579,523$ ratings of $26,629$ restaurants from $478,841$ users. We extract $10^3$ restaurants with most reviews and $10^3$ users who review most for experiments.

Before we start, we randomly choose $100$ users and formulate a binary matrix $H \in \RR^{100\times 1000}$ (stands for `history') where $H(i,k) = 1$ denotes the user $i$ has rated restaurant $k$ and $H(i,k) = 0$ denotes otherwise. We want to construct feature vectors for $10^3$ restaurants from the records of $100$ users and then use them to conduct experiments on the records of the remaining $900$ users. Then, we perform SVD on $H$ to get a $d=20$ feature vectors for each of the chosen restaurants. The remaining ratings form another binary matrix $F \in \RR^{900 \times 1000}$ (stands for `future'), which is used for online experiments.

For each of the following settings, we randomly choose $L=200$ (or $10^3$) restaurants and $u=40$ (or $200$) users. At each time step $t$, a user is selected uniformly and the learning agent recommends $K=4$ restaurants to the user. By referring the binary matrix $F$, the learning agent receives a feedback $\bC_t \in \{1,...,K,\infty\}$ and updates its statistics. The objective is to maximize the cumulative clicks of the learning agent\footnote{Since there is no universal truth about correct clustering and choosing the gap parameter $\gamma$ is quite subjective, to avoid disputes and be consistent with previous works \cite{li2016contextual}, we adopt the measure of cumulative rewards here.}. The results are shown in \cref{fig:yelp_reward}, where the vertical axis denotes the cumulative rewards and the horizontal axis is the time step $t$. From the results, the performance of our algorithm has a clear advantage over theirs.

\subsection{MovieLens Dataset}

\begin{figure}
\centering
\subfigure[$u=40,~L=200$]{
\includegraphics[width = 0.225\textwidth]{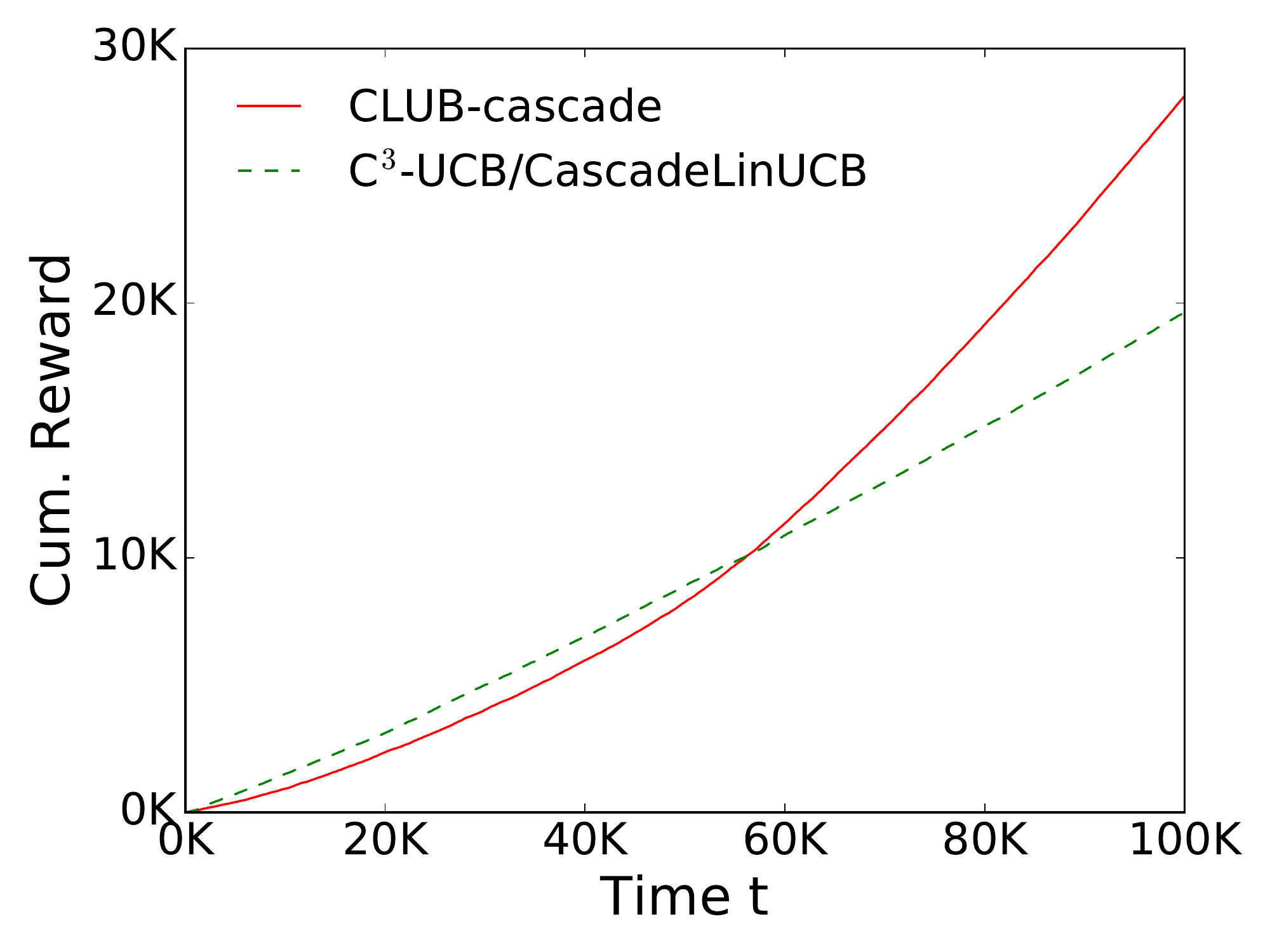}}
\subfigure[$u=40,~L=1k$]{
\includegraphics[width = 0.225\textwidth]{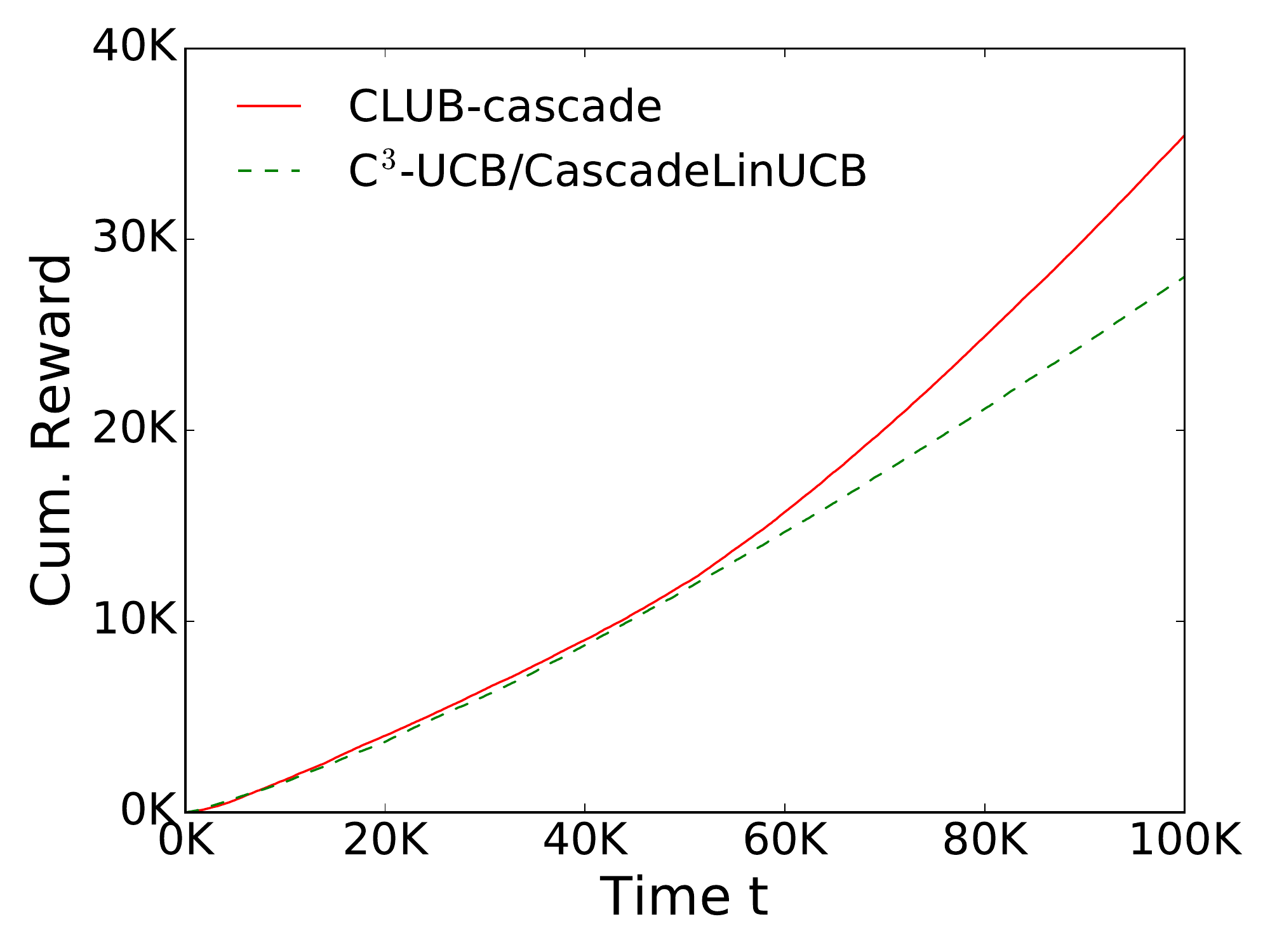}}
\subfigure[$u=200,~L=200$]{
\includegraphics[width = 0.225\textwidth]{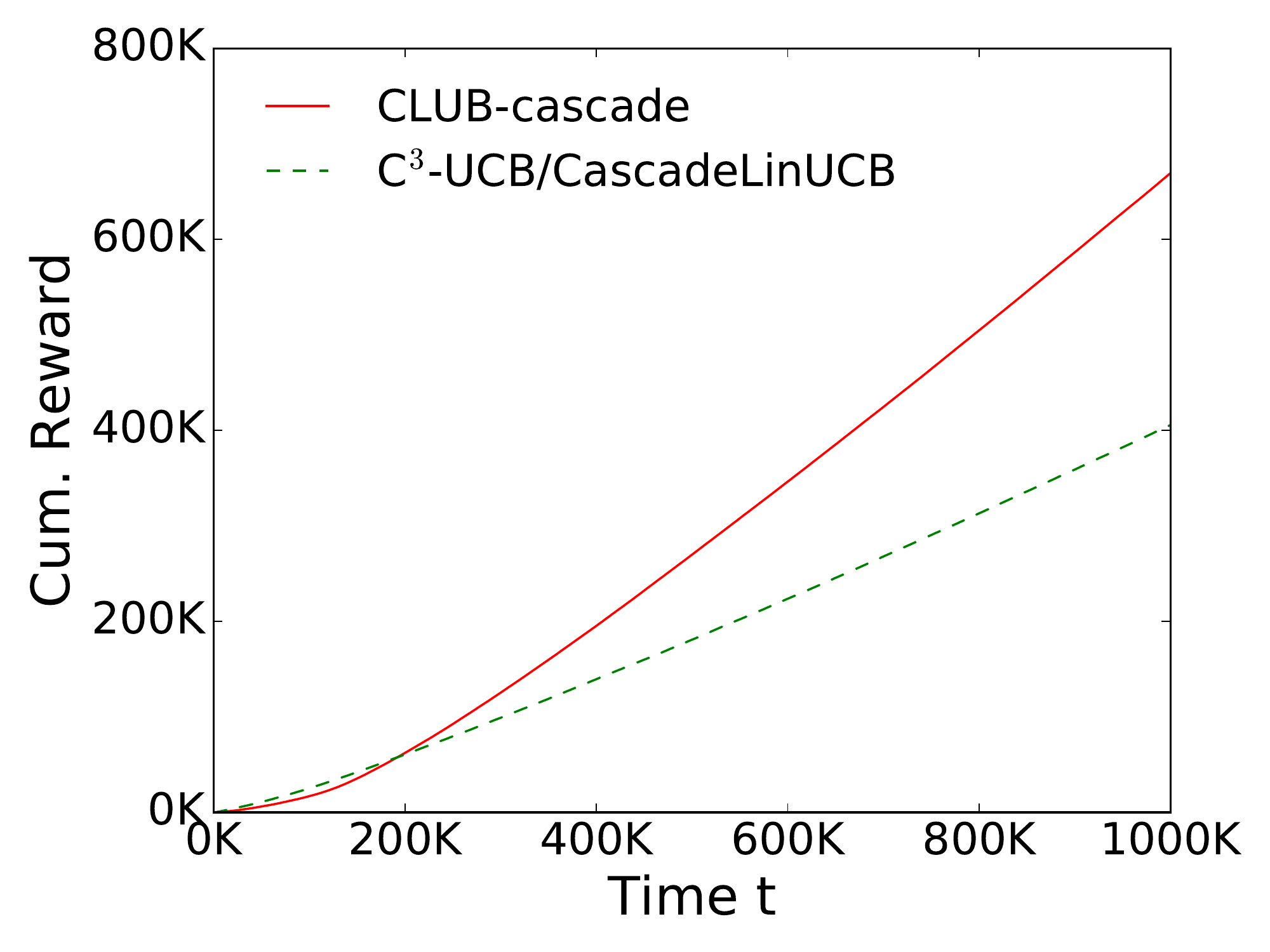}}
\subfigure[$u=200,~L=1k$]{
\includegraphics[width = 0.225\textwidth]{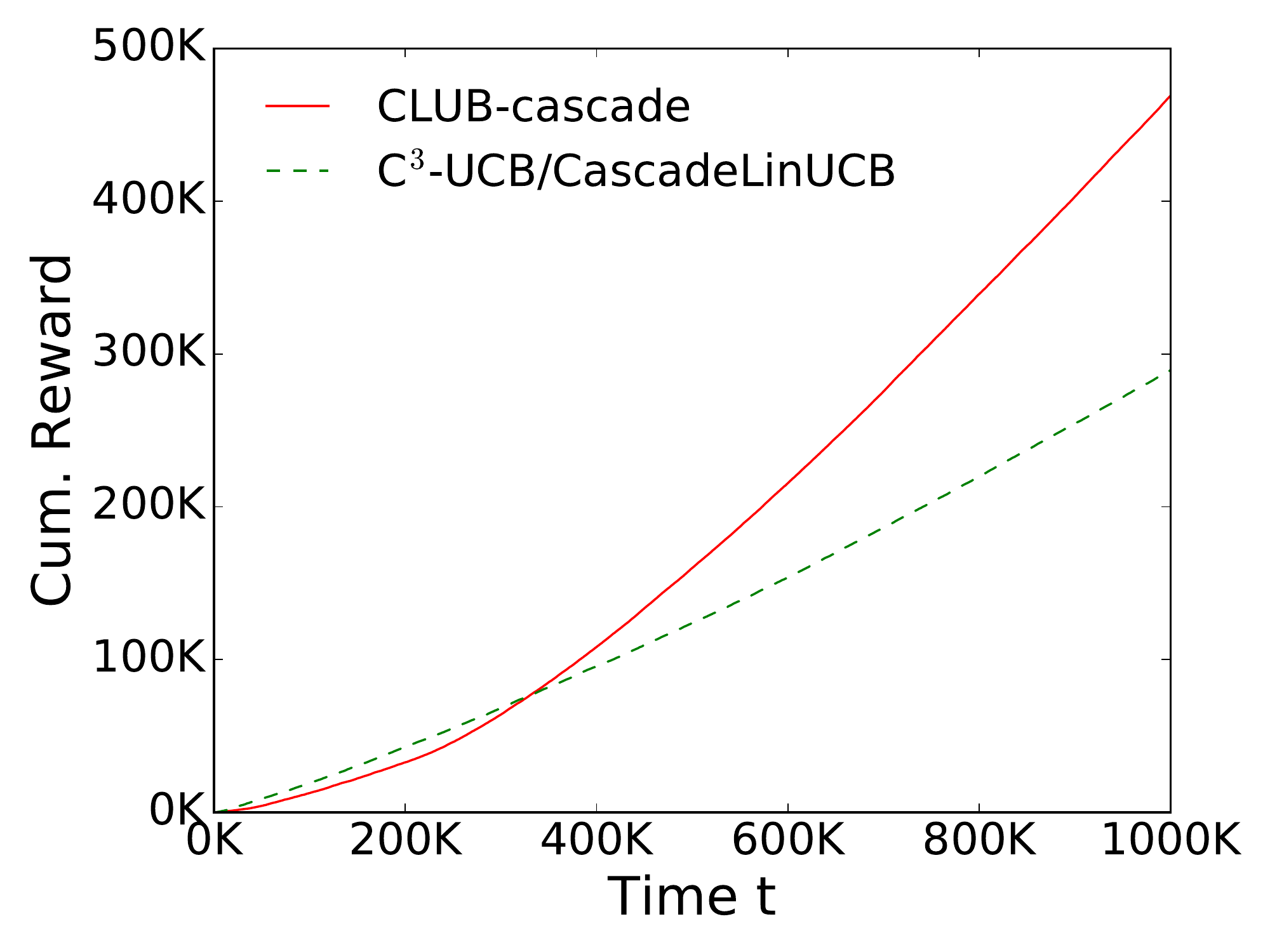}}
\caption{Cumulative clicks on MovieLens, $d=20, K=4$}
\label{fig:movie_reward}
\end{figure}

In this experiment, we compare our algorithm, CLUB-cascade, with C$^3$-UCB / CascadeLinUCB on the real dataset MovieLens \cite{harper2016movielens}. We use the processed 20m dataset\footnote{$\mathrm{https://grouplens.org/datasets/movielens/20m/}$}, in which there are $20$ million ratings for $2.7\times 10^4$ movies by $1.38\times 10^5$ users.

Since the MovieLens dataset has been processed and all users and movies have records with similar density, we randomly draw $10^3$ movies and $10^3$ users for experiments. After that, we randomly draw $100$ users from the $10^3$ users and formulate a binary matrix $H \in \RR^{100\times 1000}$, where $H(i,j) = 1$ denotes the user $i$ has rated movie $j$ and $H(i,j) = 0$ denotes otherwise. Then we perform SVD on $H$ to get a $d=20$ feature vectors for all the $10^3$ chosen movies. The records for the remaining $900$ users form another binary matrix $F \in \RR^{900 \times 1000}$, which is used for online evaluations.

For each of the four settings, we randomly choose $L=200$ (or $10^3$) movies and $u=40$ (or $200$) users. At each time step $t$, a user is selected uniformly and the learning agent recommends $K=4$ movies to the user. By referring to the binary matrix $F$, the learning agent receives a feedback $\bC_t \in \{1,...,K,\infty\}$ and updates its statistics. The objective is to maximize the cumulative clicks of the learning agent. The results are shown in \cref{fig:movie_reward}, where the vertical axis denotes the cumulative rewards and the horizontal axis is the time step $t$. From the results, the performance of our algorithm has a clear advantage over theirs.

\begin{figure}
\centering
\subfigure[Yelp]{
\includegraphics[width = 0.225\textwidth]{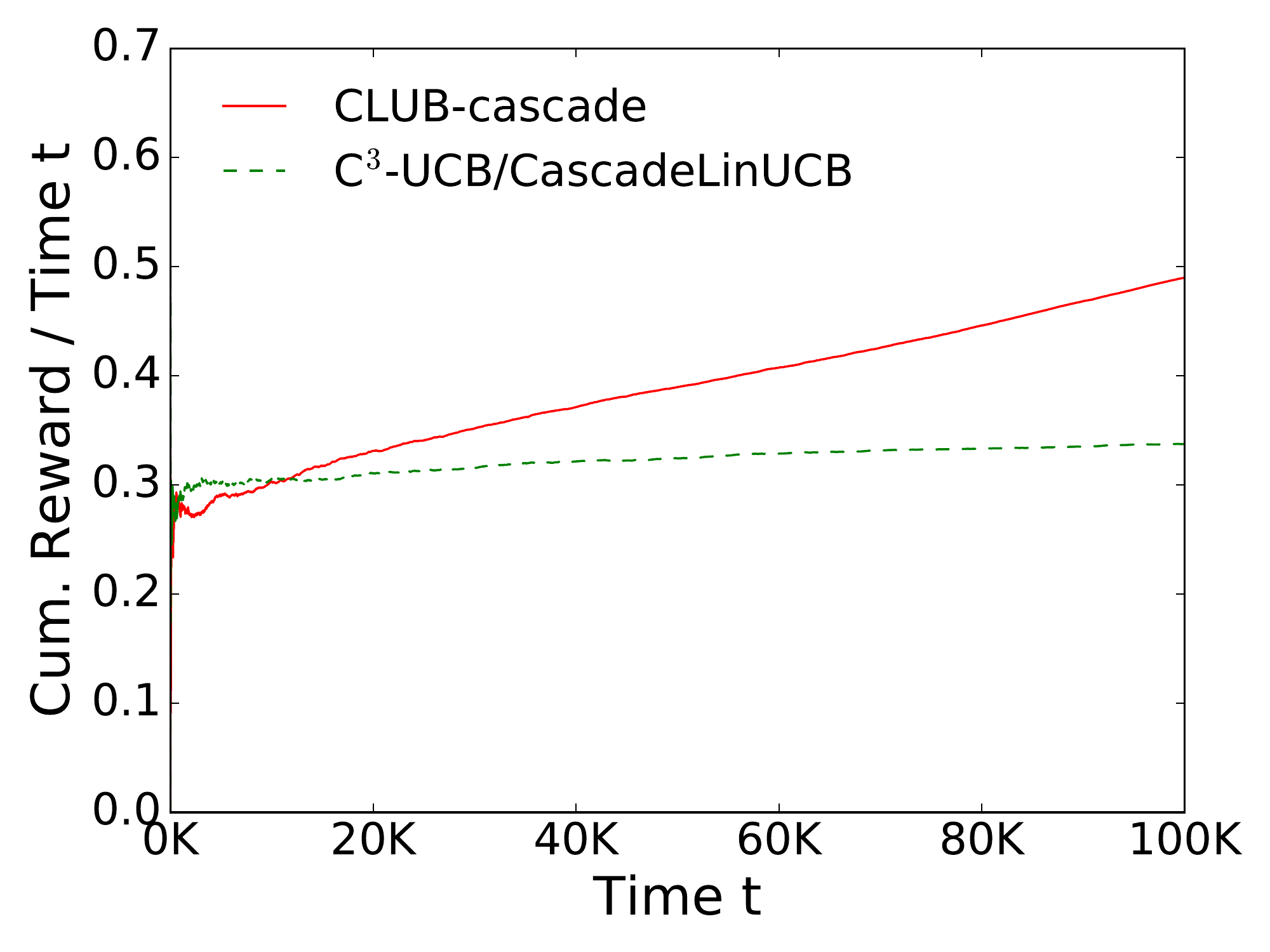}}
\subfigure[MovieLens]{
\includegraphics[width = 0.225\textwidth]{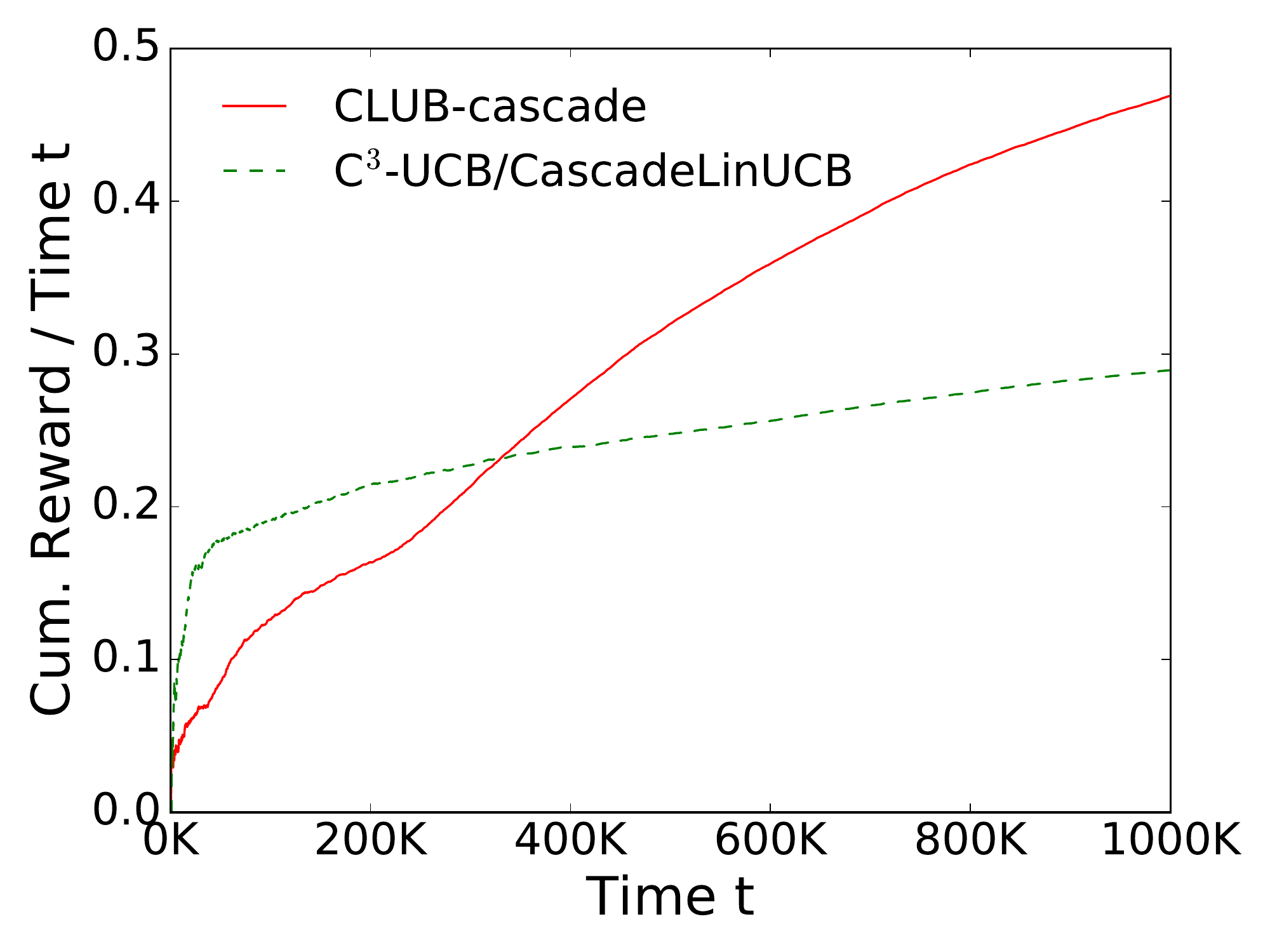}}
\caption{Comparisons of CTR on Yelp and MovieLens, $d=20, K=4, u=200, L=1000$}
\label{fig:yelp_movie_ratio}
\end{figure}

\subsection{Other comparisons}

Comparing the performances on two datasets, our algorithm seems to need more steps to obtain an obvious advantage in the MovieLens dataset. This is because the MovieLens dataset has been processed and the user-movie matrix we are dealing with is quite dense, thus users are more similar. To see this phenomenon more clearly, we draw the results on the average rewards, the cumulative rewards up to time $t$ divided by $t$, for both datasets in Figure \ref{fig:yelp_movie_ratio}. In earlier rounds, their algorithm taking all users as one will have a temporary advantage in MovieLens dataset since the users are similar. At the same time, our algorithm pays the cost of exploring clusters and starts with low average rewards. However, as the explored cluster structure becomes more and more accurate, our algorithm benefits from it and keeps a high increasing rate. As time goes by, the cost for regarding users as one is not negligible and our algorithm outperforms theirs. In most real applications, the user-item matrix would be very sparse and the users are tending to be dissimilar, resulting in a more advantaged environment for our algorithm.

\begin{figure}
\centering
\subfigure[Yelp, $u=40$]{
\includegraphics[width = 0.225\textwidth]{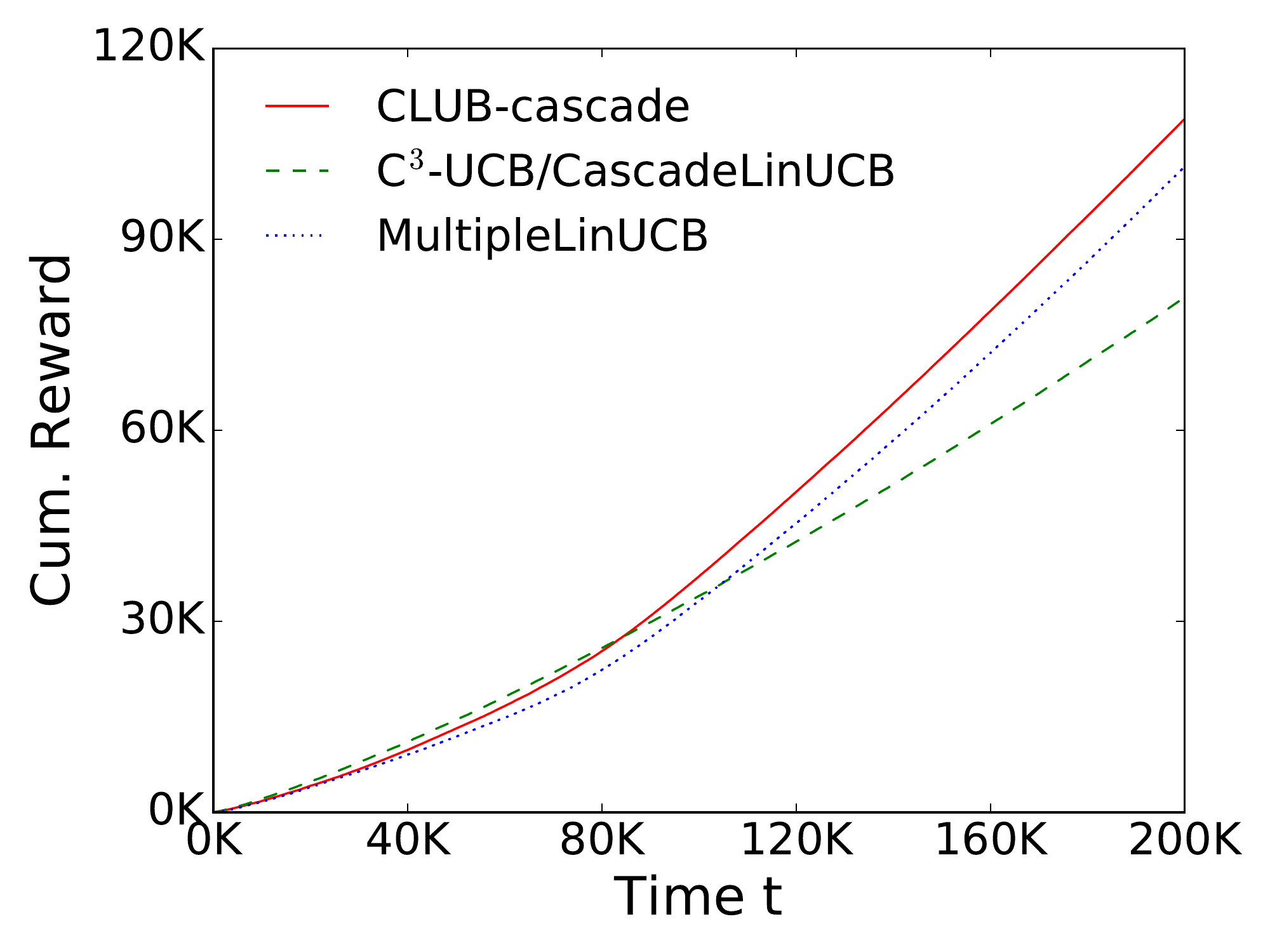}}
\subfigure[Yelp, $u=900$]{
\includegraphics[width = 0.225\textwidth]{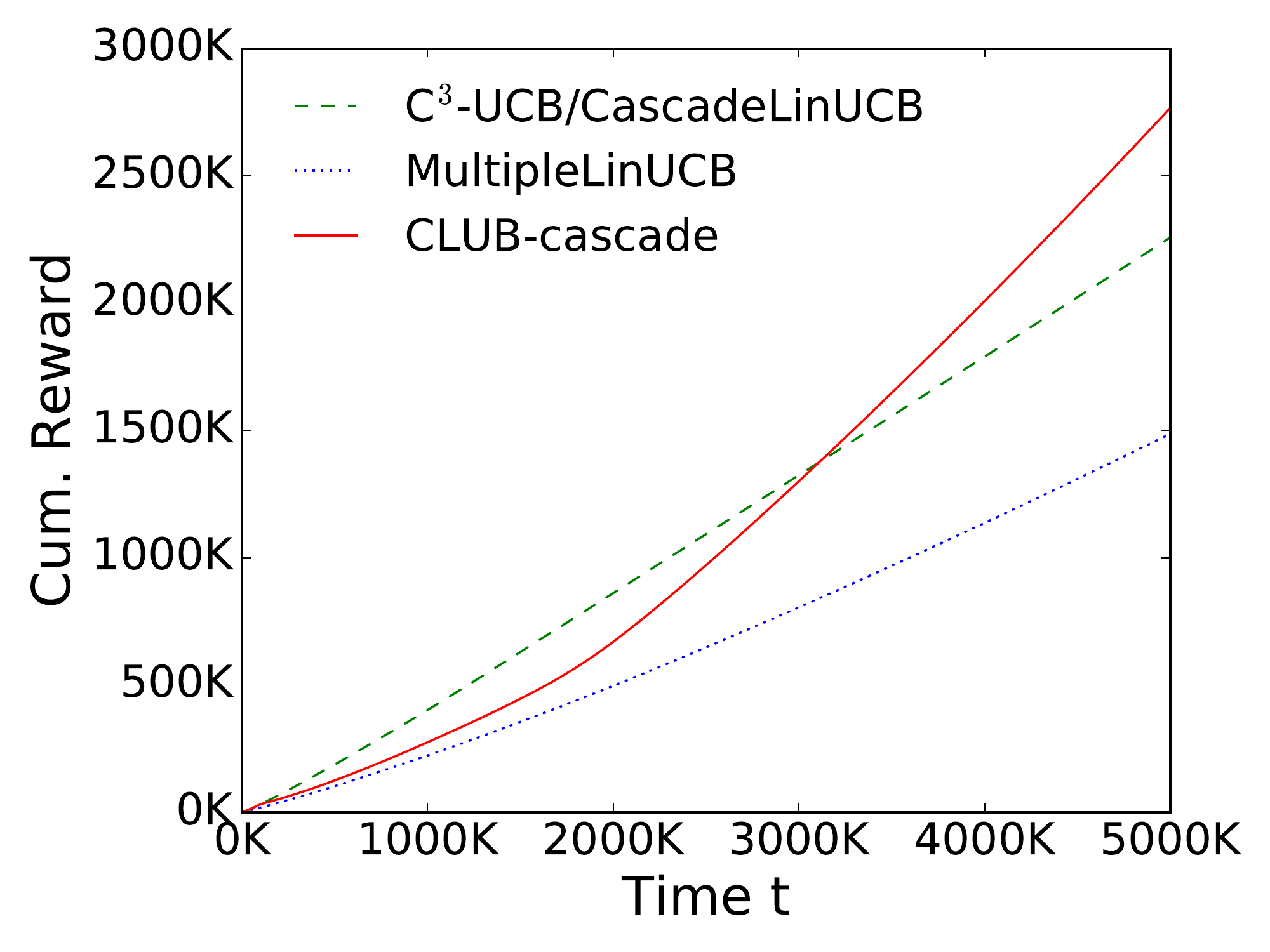}}
\subfigure[MovieLens, $u=40$]{
\includegraphics[width = 0.225\textwidth]{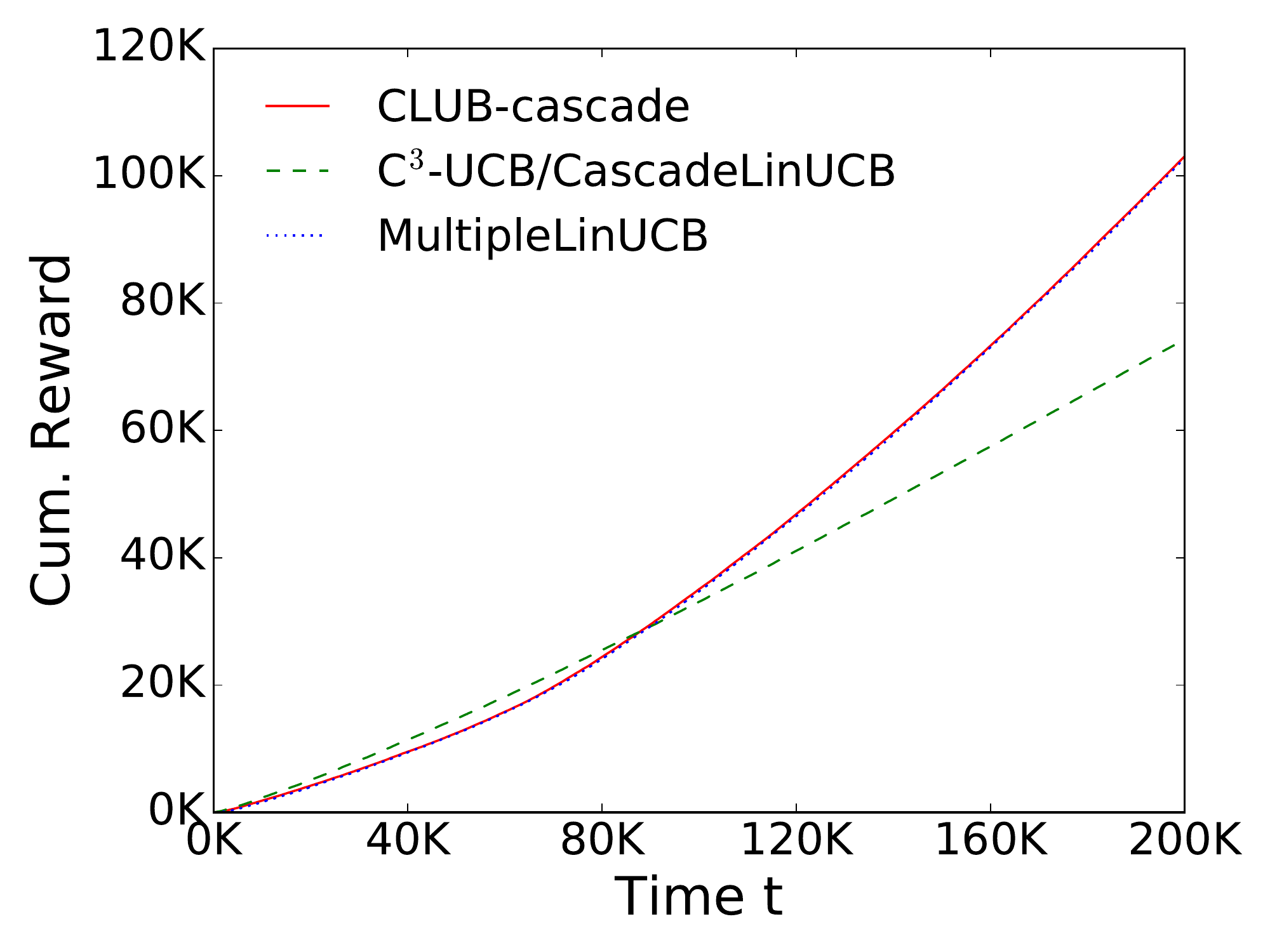}}
\subfigure[MovieLens, $u=900$]{
\includegraphics[width = 0.225\textwidth]{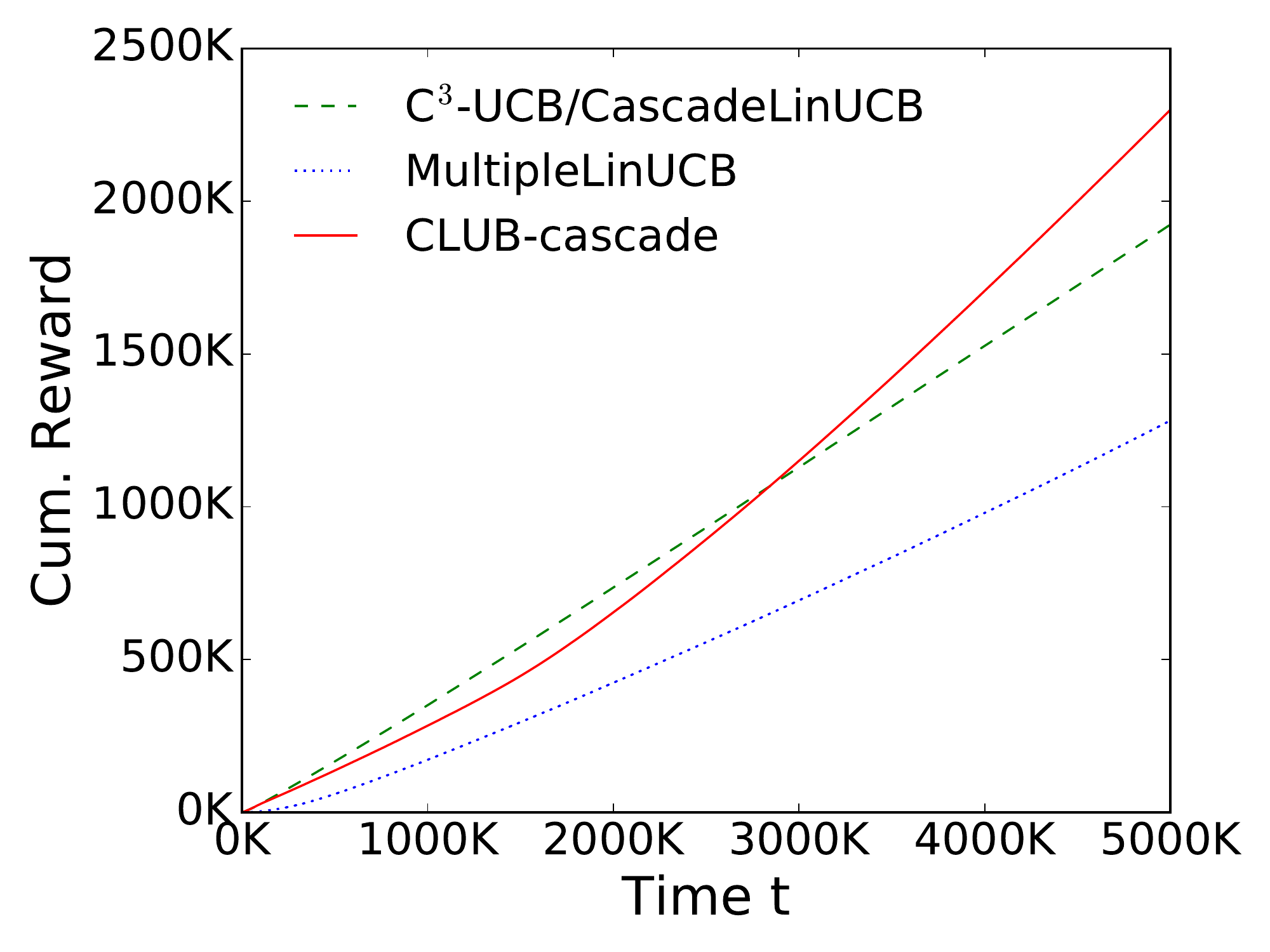}}
\caption{Comparisons with MultipleLinUCB, $d=20, K=4, L=1000$}
\label{fig:multiple}
\end{figure}

Last we show the comparisons with the algorithm MultipleLinUCB, which regards each user as an individual cluster, on both datasets. Results are shown in Figure \ref{fig:multiple}. From the figures, our algorithm has significant advantages over MultipleLinUCB when users are more. When there are many users, it takes a long time for each individual user to collect enough information and borrowing histories from other users would help improve the quality of recommendations. When there are few users, each user is able to collect enough information, thus borrowing histories from others will not help that much. Note that our algorithm has a bigger advantage on MovieLens dataset than on Yelp dataset. The reason is similar to that of the above ratio differences.

\section{6. Conclusions}
In this paper, we bring up a new problem of online clustering of contextual cascading bandits, where the algorithm has to explore the unknown cluster structure on users under a prefix feedback of the recommended item list. We propose a CLUB-cascade algorithm based on the principle of optimism in face of uncertainty and prove a cumulative regret bound, whose degenerate case improves the existing results. The experiments conducted on both synthetic and real datasets demonstrate the advantage of incorporating online clustering.

\bibliography{CCB}

\begin{thebibliography}{}

\bibitem[\protect\citeauthoryear{Abbasi-Yadkori, P{\'a}l, and
  Szepesv{\'a}ri}{2011}]{abbasi2011improved}
Abbasi-Yadkori, Y.; P{\'a}l, D.; and Szepesv{\'a}ri, C.
\newblock 2011.
\newblock Improved algorithms for linear stochastic bandits.
\newblock In {\em Advances in Neural Information Processing Systems},
  2312--2320.

\bibitem[\protect\citeauthoryear{Chuklin, Markov, and
  Rijke}{2015}]{chuklin2015click}
Chuklin, A.; Markov, I.; and Rijke, M.~d.
\newblock 2015.
\newblock Click models for web search.
\newblock {\em Synthesis Lectures on Information Concepts, Retrieval, and
  Services} 7(3):1--115.

\bibitem[\protect\citeauthoryear{Combes \bgroup et al\mbox.\egroup
  }{2015}]{combes2015learning}
Combes, R.; Magureanu, S.; Proutiere, A.; and Laroche, C.
\newblock 2015.
\newblock Learning to rank: Regret lower bounds and efficient algorithms.
\newblock {\em ACM SIGMETRICS Performance Evaluation Review} 43(1):231--244.

\bibitem[\protect\citeauthoryear{Craswell \bgroup et al\mbox.\egroup
  }{2008}]{craswell2008experimental}
Craswell, N.; Zoeter, O.; Taylor, M.; and Ramsey, B.
\newblock 2008.
\newblock An experimental comparison of click position-bias models.
\newblock In {\em Proceedings of the 2008 International Conference on Web
  Search and Data Mining},  87--94.
\newblock ACM.

\bibitem[\protect\citeauthoryear{Dani, Hayes, and
  Kakade}{2008}]{dani2008stochastic}
Dani, V.; Hayes, T.~P.; and Kakade, S.~M.
\newblock 2008.
\newblock Stochastic linear optimization under bandit feedback.
\newblock In {\em COLT},  355--366.

\bibitem[\protect\citeauthoryear{Filippi \bgroup et al\mbox.\egroup
  }{2010}]{filippi2010parametric}
Filippi, S.; Cappe, O.; Garivier, A.; and Szepesv{\'a}ri, C.
\newblock 2010.
\newblock Parametric bandits: The generalized linear case.
\newblock In {\em Advances in Neural Information Processing Systems},
  586--594.

\bibitem[\protect\citeauthoryear{Gentile \bgroup et al\mbox.\egroup
  }{2017}]{gentile2017context}
Gentile, C.; Li, S.; Kar, P.; Karatzoglou, A.; Zappella, G.; and Etrue, E.
\newblock 2017.
\newblock On context-dependent clustering of bandits.
\newblock In {\em International Conference on Machine Learning},  1253--1262.

\bibitem[\protect\citeauthoryear{Gentile, Li, and
  Zappella}{2014}]{gentile2014online}
Gentile, C.; Li, S.; and Zappella, G.
\newblock 2014.
\newblock Online clustering of bandits.
\newblock In {\em Proceedings of the 31st International Conference on Machine
  Learning},  757--765.

\bibitem[\protect\citeauthoryear{Harper and
  Konstan}{2016}]{harper2016movielens}
Harper, F.~M., and Konstan, J.~A.
\newblock 2016.
\newblock The movielens datasets: History and context.
\newblock {\em ACM Transactions on Interactive Intelligent Systems (TiiS)}
  5(4):19.

\bibitem[\protect\citeauthoryear{Katariya \bgroup et al\mbox.\egroup
  }{2016}]{katariya2016dcm}
Katariya, S.; Kveton, B.; Szepesvari, C.; and Wen, Z.
\newblock 2016.
\newblock Dcm bandits: Learning to rank with multiple clicks.
\newblock In {\em Proceedings of The 33rd International Conference on Machine
  Learning},  1215--1224.

\bibitem[\protect\citeauthoryear{Kveton \bgroup et al\mbox.\egroup
  }{2015a}]{kveton2015cascading}
Kveton, B.; Szepesv{\'a}ri, C.; Wen, Z.; and Ashkan, A.
\newblock 2015a.
\newblock Cascading bandits: learning to rank in the cascade model.
\newblock In {\em Proceedings of the 32nd International Conference on Machine
  Learning}.

\bibitem[\protect\citeauthoryear{Kveton \bgroup et al\mbox.\egroup
  }{2015b}]{kveton2015combinatorial}
Kveton, B.; Wen, Z.; Ashkan, A.; and Szepesvari, C.
\newblock 2015b.
\newblock Combinatorial cascading bandits.
\newblock In {\em Advances in Neural Information Processing Systems},
  1450--1458.

\bibitem[\protect\citeauthoryear{Lattimore and
  Szepesvari}{2017}]{lattimore2017end}
Lattimore, T., and Szepesvari, C.
\newblock 2017.
\newblock The end of optimism? an asymptotic analysis of finite-armed linear
  bandits.
\newblock In {\em Artificial Intelligence and Statistics},  728--737.

\bibitem[\protect\citeauthoryear{Li \bgroup et al\mbox.\egroup
  }{2016}]{li2016contextual}
Li, S.; Wang, B.; Zhang, S.; and Chen, W.
\newblock 2016.
\newblock Contextual combinatorial cascading bandits.
\newblock In {\em Proceedings of The 33rd International Conference on Machine
  Learning},  1245--1253.

\bibitem[\protect\citeauthoryear{Li, Karatzoglou, and
  Gentile}{2016}]{li2016collaborative}
Li, S.; Karatzoglou, A.; and Gentile, C.
\newblock 2016.
\newblock Collaborative filtering bandits.
\newblock In {\em Proceedings of the 39th International ACM SIGIR conference on
  Research and Development in Information Retrieval},  539--548.
\newblock ACM.

\bibitem[\protect\citeauthoryear{Li, Lu, and Zhou}{2017}]{li2017provably}
Li, L.; Lu, Y.; and Zhou, D.
\newblock 2017.
\newblock Provably optimal algorithms for generalized linear contextual
  bandits.
\newblock In {\em International Conference on Machine Learning},  2071--2080.

\bibitem[\protect\citeauthoryear{Tropp}{2011}]{tropp2011freedman}
Tropp, J.~A.
\newblock 2011.
\newblock Freedman’s inequality for matrix martingales.
\newblock {\em Electron. Commun. Probab} 16:262--270.

\bibitem[\protect\citeauthoryear{Zoghi \bgroup et al\mbox.\egroup
  }{2017}]{zoghi2017online}
Zoghi, M.; Tunys, T.; Ghavamzadeh, M.; Kveton, B.; Szepesvari, C.; and Wen, Z.
\newblock 2017.
\newblock Online learning to rank in stochastic click models.
\newblock In {\em International Conference on Machine Learning},  4199--4208.

\bibitem[\protect\citeauthoryear{Zong \bgroup et al\mbox.\egroup
  }{2016}]{zong2016cascading}
Zong, S.; Ni, H.; Sung, K.; Ke, N.~R.; Wen, Z.; and Kveton, B.
\newblock 2016.
\newblock Cascading bandits for large-scale recommendation problems.
\newblock In {\em the 32nd Conference on Uncertainty in Artificial
  Intelligence}.

\end{thebibliography}
\bibliographystyle{aaai}

\newpage
\appendix
\onecolumn
\section{A. Proof of Theorem \ref{thm:main}}

Similarly in \cite{kveton2015cascading}, we define a permutation of $X_t^\ast = (x_{t,1}^\ast,\ldots,x_{t,K}^\ast)$ to satisfy that if $\bx_{t,k} \in X_t^\ast$, then set $x_{t,k}^\ast = \bx_{t,k}$, and other optimal items are positioned randomly. Under this arrangement,
$$
\forall k \in [K], \qquad y_t(x_{t,k}^\ast) \ge y_t(\bx_{t,k}), \qquad \bU_t(x_{t,k}^\ast) \le \bU_t(\bx_{t,k}).
$$

First the instantaneous regret for round $t$ is
\begin{align*}
&\quad~\EE_t[R(\bA_t, \by_t)] = \EE_t\left[ \left( 1 - \prod_{k=1}^{K}(1 - \by_t(x_{t,k}^\ast)) \right) - \left( 1 -  \prod_{k=1}^{K}(1 - \by_t(\bx_{t,k})) \right)  \right]\\
&= \EE_t\left[\prod_{k=1}^{K}(1 - \by_t(\bx_{t,k})) - \prod_{k=1}^{K}(1 - \by_t(x_{t,k}^\ast))\right]\\
&= \EE_t\left[ \sum_{k=1}^{K} \left( \prod_{\ell=1}^{k-1} (1 - \by_t(\bx_{t,\ell})) \right) \left[ (1 - \by_t(\bx_{t,k})) - (1 - \by_t(x_{t,k}^\ast)) \right] \left( \prod_{\ell=k+1}^{K} (1 - \by_t(x_{t,\ell}^\ast)) \right) \right] \numberthis \label{eq:pfReg1}\\
&=\EE_t\left[ \sum_{k=1}^{K} \left( \prod_{\ell=1}^{k-1} (1 - \by_t(\bx_{t,\ell})) \right) \left[ \by_t(x_{t,k}^\ast) - \by_t(\bx_{t,k}) \right] \left( \prod_{\ell=k+1}^{K} (1 - \by_t(x_{t,\ell}^\ast)) \right) \right]\\
&\le \EE_t\left[ \sum_{k=1}^{K} \prod_{\ell=1}^{k-1}(1 - \by_t(\bx_{t,\ell})) [ \by_t(x_{t,k}^\ast) - \by_t(\bx_{t,k}) ] \right] \numberthis \label{eq:pfReg2}\\
\end{align*}
where \eqref{eq:pfReg1} is by \cite[Lemma B.1]{kveton2015cascading} and \eqref{eq:pfReg2} is by $\by(\cdot) \in \{0,1\}$. 

Let
\begin{align*}
\bbeta(T,\delta) = \sqrt{d\ln\left(1+T/(\lambda d)\right) + 2\ln\frac{1}{\delta}} + \sqrt{\lambda}
\end{align*}
Define events
\begin{align*}
\cE &= \{ \text{the clusters are correct for all } t > T_0\},\\
\cF_{j}(\delta) &= \left\{ \norm{\theta_j - \hat{\btheta}_{I_j,t}}_{\bM_{I_j,t}} \le \bbeta(T_{I,t-1},\delta), ~~\forall t \right\},~~\forall j\in[m],\\
\cG_{t,k} &= \{ \text{item } \bx_{t,k} \text{ is examined at time } t \}, \forall t\ge 1,k\in[K],
\end{align*}
where $T_0$ will be specified later. Note that $\bOne\{\cG_{t,k}\} = \prod_{\ell=1}^{k-1} \left(1 - \by_t(\bx_{t,\ell})\right)$. Then
\begin{align*}
R(T) &= \EE\left[\sum_{t=1}^T \EE_t\left[R(\bA_t, \by_t)\right]\right] = \EE\left[\sum_{t=1}^{T_0} \EE_t\left[R(\bA_t, \by_t)\right]\right] + \EE\left[\sum_{t=T_0+1}^T \EE_t\left[R(\bA_t, \by_t)\right]\right]\\
&\le T_0 + \EE\left[\sum_{t=T_0+1}^T \EE_t\left[R(\bA_t, \by_t)\right]\right]\\
&= T_0 + \PP\left(\cE,\cF_1(\delta),\ldots,\cF_m(\delta) \right) \EE\left[\sum_{t=T_0+1}^T \EE_t\left[R(\bA_t, \by_t)\right]\bigg| \cE, \cF_1(\delta), \ldots,\cF_m(\delta) \right] \\
& \qquad \qquad \qquad + \PP\left(\bigcup_{j=1}^m \cF_j(\delta)^c \bigcup \cE^c \right) \EE\left[\sum_{t=T_0+1}^T \EE_t\left[R(\bA_t, \by_t)\right]\bigg| \bigcup_{j=1}^m \cF_j^c(\delta) \bigcup \cE^c \right] \\
&\le T_0 + \EE\left[\sum_{t=T_0+1}^T \EE_t\left[R(\bA_t, \by_t)\right]\bigg| \cE, \cF_1(\delta), \ldots, \cF_m(\delta) \right] + (T-T_0) \left[\PP(\cE^c) + \sum_{j=1}^m \PP(\cF_j^c(\delta)) \right]
\end{align*}
And
\begin{align*}
&\quad ~\EE\left[\sum_{t=T_0+1}^T \EE_t\left[R(\bA_t, \by_t)\right]\bigg| \cE, \cF_1(\delta), \ldots, \cF_m(\delta) \right]\\
&\le \EE \left[ \sum_{t=T_0+1}^{T} \sum_{k=1}^{K} \EE_t\left[ \bOne\{ \cG_{t,k} \} \right] \left[ y_{j_t}(x_{t,k}^\ast) - y_{j_t}(\bx_{t,k}) \right] \bigg| \cE, \cF_1(\delta), \ldots, \cF_m(\delta) \right] \\
&\le \EE \left[ \sum_{t=T_0+1}^{T} \sum_{k=1}^{K} \EE_t\left[ \bOne\{ \cG_{t,k} \} \right] \cdot 2\bbeta(T_{I_{j_t},t-1},\delta) \norm{\bx_k^t}_{\bM_{I_{j_t},t}^{-1}} \bigg| \cE, \cF_1(\delta),\ldots,\cF_m(\delta) \right] \numberthis \label{eq:pfReg3}\\
&\le \EE \left[ \sum_{t=T_0+1}^{T} 2\bbeta(T_{I_{j_t},t-1},\delta) \sum_{k=1}^{\bK_t} \norm{\bx_k^t}_{\bM_{I_{j_t},t}^{-1}} \bigg| \cE, \cF_1(\delta),\ldots,\cF_m(\delta) \right] \\
&\le 2\EE \left[ \sum_{j=1}^m \bbeta(T_{I_j,T-1},\delta) \sum_{\substack{T_0 < t \le T \\ i_t \in I_j}}  \sum_{k=1}^{\bK_t} \norm{\bx_k^t}_{\bM_{I_j,t}^{-1}} \bigg| \cE, \cF_1(\delta),\ldots,\cF_m(\delta) \right] \\
&\le 2 \sum_{j=1}^m \left(\sqrt{d\ln\left(1+\frac{T_{I_j}}{\lambda d}\right) + 2\ln\frac{1}{\delta}} + \sqrt{\lambda}\right) \sqrt{2d T_{I_j} K\ln\left(1+\frac{T_{I_j} K}{\lambda d}\right)} \numberthis \label{eq:pfReg4}\\
&\le 2 \left(\sqrt{d\ln\left(1+\frac{T}{\lambda d}\right) + 2\ln\frac{1}{\delta}} + \sqrt{\lambda}\right) \sqrt{2dmKT\ln\left(1+\frac{KT}{\lambda d}\right)}
\end{align*}
where \eqref{eq:pfReg3} is by
\begin{align*}
y_{j_t}(\bx_{t,k}) \le y_{j_t}(x_{t,k}^\ast) &\le \bU_t(x_{t,k}^\ast) \le \bU_t(\bx_{t,k}) \le \hat{\btheta}_{t-1}^{\top}\bx_{t,k} + \bbeta(T_{I_{j_t},t-1},\delta) \norm{\bx_{t,k}}_{\bM_{I_{j_t},t}^{-1}} \\
&\le y_{j_t}(\bx_{t,k}) + 2\bbeta(T_{I_{j_t},t-1},\delta) \norm{\bx_{t,k}}_{\bM_{I_{j_t},t}^{-1}}
\end{align*}
and \eqref{eq:pfReg4} is by \eqref{eq:selfNormSum}. Next is to give $T_0$ and bound $\PP(\cE^c), \PP(\cF_j^c)$.

Define events for any $i\in [u]$,
\begin{align*}
\cB_{1i}(\delta) &= \left\{ \norm{\hat{\btheta}_{i,t} - \theta_{j(i)}}_{\bM_{i,t-1}} \le \bbeta(T_{i,t}, \delta), \text{  for all  } t \ge 1\right\},\\
\cB_{2i}(\delta) &= \left\{ \lambda_{\min}(\cS_{i,t}) \ge T_{i,t} \lambda_x/8, \text{  for all  } T_{i,t} \ge \frac{1024}{\lambda_x^2}\ln\frac{512d}{\lambda_x^2 \delta}
\right\}. 
\end{align*}

The high probability property of these events is postponed to next section. Specifically, $\PP(\cF_j(\delta)^c) \le \delta$ and $\PP(\cB_{1i}(\delta)^c) \le \delta$ by Lemma \ref{lem:thetaEstimate} and Lemma \ref{lem:detMnUB}; $\PP(\cB_{2i}(\delta)^c) \le \delta$ by Claim 1 of \cite{gentile2014online}, Lemma \ref{lem:lambdaMinOfM}, \ref{lem:sumBerGeq} and the assumption of user uniformness.

Under the event $\cB_{1i}(\delta/(4u)), \cB_{2i}(\delta/(4u))$, for all $i\in[u]$, which holds with probability at least $1-3\delta/4$, then
\begin{align*}
\norm{\hat{\btheta}_{i,t} - \theta_{j(i)}}_2 
\le \frac{\norm{\hat{\btheta}_{i,t} - \theta_{j(i)}}_{\bM_{i,t}}}{\sqrt{\lambda_{\min}(\bM_{i,t})} } 
\le \frac{\bbeta(T_{i,t},\delta/4u)}{\sqrt{\lambda + \lambda_{\min}(\bS_{i,t})} } 
\le \frac{\sqrt{d\ln\left(1+\frac{T_{i,t}}{\lambda d}\right) + 2\ln\frac{4u}{\delta} } + \sqrt{\lambda}}{ \sqrt{\lambda + T_{i,t}\lambda_x/8} } 
< \frac{\gamma}{2}\,,
\end{align*}
where the last inequality is valid when 
\begin{align*}
T_{i,t} \ge \frac{512d}{\gamma^2\lambda_x}\ln\frac{4u}{\delta}
\end{align*}
and the proof is postponed to Lemma \ref{lem:tForLeGamma}. 

In this case, if the condition of deleting edge in the \cref{alg:ClubCascade} (Line \ref{alg:clubcascade:delete}), we could obtain $\norm{\theta_{j(i)} - \theta_{j(\ell)}} > 0$. By the assumption of cluster regularity, $\norm{\theta_{j(i)} - \theta_{j(\ell)}} \ge \gamma$, thus user $i$ and user $\ell$ belong to different clusters. On the other hand, if user $i$ and user $\ell$ belong to different clusters, then the $\norm{\cdot}_2$ confidence radius for the weight vectors tends to $0$, thus the required condition in \cref{alg:ClubCascade} (Line \ref{alg:clubcascade:delete}) will be satisfied when $T_{i,t}$ and $T_{\ell,t}$ satisfies the above condition. If this holds for all users, the obtained clusters will be correct.

Combining with the condition in $\cB_{2i}$, it is required that
\begin{align*}
T_{i,t} \ge \max\left\{ \frac{512d}{\gamma^2\lambda_x}\ln\frac{4u}{\delta}, \quad \frac{256}{\lambda_x^2}\ln\frac{128d}{\lambda_x^2 \delta} \right\}, \text{    for all } i\in[u]\,,
\end{align*}
which can be satisfied when
\begin{align*}
t \ge 16u\ln\frac{4uT}{\delta} + 4u \max\left\{ \frac{512d}{\gamma^2\lambda_x}\ln\frac{4u}{\delta}, ~~\frac{256}{\lambda_x^2}\ln\frac{128d}{\lambda_x^2 \delta} \right\} =: T_0(\delta)
\end{align*}
with probability at least $1-\delta/4$.

Under all the above conditions, the clusters are correctly partitioned, that is, the event $\cE$ holds with probability at least $1-3\delta/4$. Next take events $\cF_j(\delta/(4m))$ and $\delta = 1/T$. Thus the cumulative regret satisfies
\begin{equation}
\label{eq:regret1}
\begin{split}
R(T) \le &2 \left(\sqrt{d\ln\left(1+\frac{T}{\lambda d}\right) + 2\ln(4mT)} + \sqrt{\lambda}\right) \sqrt{2dmKT\ln\left(1+\frac{TK}{\lambda d}\right)} + O\left( u\left(\frac{d}{\gamma^2 \lambda_x} + \frac{1}{\lambda_x^2}\right) \ln(T) \right)\,.
\end{split}
\end{equation}

\subsection{Proof of Corollary \ref{cor:m=1}}

In this special case that all users belong to only one cluster, which is linear cascading bandit setting, the exploration rounds on clusters will not be needed. In the decomposition formula for cumulative regrets $R(T)$, the initialized complete graph represents the correct clustering, thus $T_0 = 0$. Then only the first line of regret bound in Eq. \eqref{eq:regret1} is needed.

\section{B. Proof of Theorem \ref{thm:glm}}
\label{sec:pfGLM}

Let 
$$
g_{I,t}(\theta) = \sum_{s=1}^{t-1} \bOne\{i_s \in I\} \sum_{k=1}^{\bK_s} \mu(\theta^\top \bx_{s,k}) \bx_{s,k}.
$$
Recall that $\hat{\btheta}_{I,t}$ is the unique solution of
$$
\sum_{s=1}^{t} \bOne\{i_s \in I\} \sum_{k=1}^{\bK_s} \left( \by_{s,k} - \mu(\theta^\top \bx_{s,k}) \right) \bx_{s,k} = 0,
$$
Then $g_{I,t}(\hat{\btheta}_{I,t}) = \sum_{s=1}^{t} \bOne\{i_s \in I\} \sum_{k=1}^{\bK_s} \by_{s,k} \bx_{s,k}$. Fix an $I \subset I_j$,
$$
g_{I,t}(\hat{\btheta}_{I,t}) - g_{I,t}(\theta_j) = \sum_{s=1}^{t} \bOne\{i_s \in I\} \sum_{k=1}^{\bK_s} (\by_{s,k} - \mu(\theta_j^\top \bx_{s,k})) \bx_{s,k}.
$$
Then by \cite[Theorem 1]{abbasi2011improved} and Lemma \ref{lem:detMnUB}, with probability at least $1-\delta$, for all $t\ge 0$,
$$
\norm{g_{I,t}(\hat{\btheta}_{I,t}) - g_{I,t}(\theta_j)}_{S_{I,t}^{-1}}^2 \le T_{I,s} \lambda_{\min}(S_{I,t})^{-1} + d \ln\frac{T_{I,t}}{d} + 2 \ln\frac{1}{\delta}
$$
for some $s \le t$ (will be clarified later) with $S_{I,s}$ invertible. Also note that 
$$
g_{I,t}(\hat{\btheta}_{I,t}) - g_{I,t}(\theta_j) = \sum_{s=1}^{t} \bOne\{i_s \in I\} \sum_{k=1}^{\bK_s} (\mu(\hat{\btheta}_{I,t}^\top \bx_{s,k}) - \mu(\theta_j^\top \bx_{s,k})) \bx_{s,k} = G_{I,t} (\hat{\btheta}_{I,t} - \theta_j)
$$
and by the property of Lipschitz and first derivative for $\mu$, $G_t \succeq c_{\mu} S_{I,t}$, then
$$
c_{\mu}^2 \norm{\hat{\btheta}_{I,t} - \theta_j}_{S_{I,t}}^2 \le T_{I,s} \lambda_{\min}(S_{I,t})^{-1} + d \ln\frac{T_{I,t}}{d} + 2 \ln\frac{1}{\delta}.
$$

The proof for the generalized linear rewards is similar to that in linear case. We only list different parts here. The event $\cF_{j}(\delta)$, $\bbeta$ and $\cB_{1i}(\delta)$ will be modified as
\begin{align*}
\bbeta(T,\delta) &= \frac{1}{c_{\mu}} \sqrt{ \frac{8}{\lambda_x} + d \ln\frac{T}{d} + 2 \ln\frac{1}{\delta}},\\
\cF_{j}(\delta) &= \left\{ \norm{\hat{\btheta}_{I_j,t} - \theta_j }_{S_{I_j,t}} \le \bbeta(T_{I,t-1},\delta), ~~\forall t \right\},~~\forall j\in[m],\\
\cB_{1i}(\delta) &= \left\{ \norm{\hat{\btheta}_{i,t} - \theta_{j(i)}}_{S_{i,t-1}} \le \bbeta(T_{i,t}, \delta), \text{  for all  } t \ge 1\right\}.
\end{align*}
And the \eqref{eq:pfReg3} is modified according to
\begin{align*}
&\abs{ y_{j_t}(x_{t,k}^\ast) - y_{j_t}(\bx_{t,k}) } = \abs{ \mu(\theta_{j_t}^{\top} x_{t,k}^\ast) - \mu( \theta_{j_t}^{\top}\bx_{t,k} ) } \le \kappa_\mu \norm{ \theta_{j_t}^{\top} x_{t,k}^\ast - \theta_{j_t}^{\top}\bx_{t,k} } \\
\le~ &2\kappa_\mu \bbeta(T_{I_{j_t},t-1},\delta) \norm{\bx_{t,k}}_{S_{I_{j_t},t}^{-1}},
\end{align*}
where the last inequality is by the same proof in linear case.

For each user $i$, for all $T_{i,t} \ge \frac{1024}{\lambda_x^2}\ln\frac{512d}{\lambda_x^2 \delta}$, $\lambda_{\min}(\cS_{i,t}) \ge T_{i,t} \lambda_x/8$ with high probability. Thus with high probability, when $t \ge T_0(\delta)$,
\begin{align*}
\norm{\hat{\btheta}_{i,t} - \theta_{j(i)}}_{S_{I,t}} &\le \frac{1}{c_{\mu}} \sqrt{ \frac{8}{\lambda_x} + d \ln\frac{T_{i,t}}{d} + 2 \ln\frac{1}{\delta}}\\
\norm{\hat{\btheta}_{i,t} - \theta_{j(i)}} &\le \frac{\sqrt{ \frac{8}{\lambda_x} + d \ln\frac{T_{i,t}}{d} + 2 \ln\frac{1}{\delta}} }{c_{\mu} \sqrt{\lambda_x T_{i,t}/8}}.
\end{align*}
Then with high probability, $\norm{\hat{\btheta}_{i,t} - \theta_j} \le 1$, thus $\norm{\hat{\btheta}_{i,t}} \le 2$. Also let $\alpha = \frac{16\sqrt{d}}{\lambda_x c_{\mu}}$. Similar properties could be derived for $\hat{\btheta}_{I_j,t}$. Thus the proof is finished.

\section{C. Technical Lemmas}

\begin{lemma} \label{lem:detMnUB}
Let $M_n = M + \sum_{t=1}^n x_t x_t^{\top}$, where $M \in \RR^{d \times d}$ is a strictly positive definite matrix and $\{x_t\}_{t=1}^n \subset \RR^d$ is a set of $d$-dimensional column vectors. Then $\det(M_n)$ is increasing in $n$ and
\begin{equation}
\det(M_n) \leq \frac{1}{d^d} \left(\trace(M) + \sum_{t=1}^n \norm{x_t}_2^2\right)^d.
\end{equation}
Furthermore, if $M = \lambda I, \lambda >0$, and $\norm{x_t} \le L, \forall t\ge 1$, then
\begin{equation}
\det(M_n) \leq (\lambda + nL^2/d)^d.
\end{equation}
\end{lemma}
\begin{proof}
Let $\lambda_1,\ldots,\lambda_d$ be the eigenvalues of $M_n$. By the inequality between arithmetric and geometric means,
\begin{align*}
\det(M_n) &= \prod_{i=1}^d \lambda_i \le \left(\frac{\lambda_1 +\ldots+\lambda_d}{d} \right)^d = (\trace(M_n)/d)^d \\
&= \left( \frac{1}{d} \left( \trace(M) + \sum_{t=1}^n \trace(x_t x_t^{\top}) \right) \right)^d = \left( \frac{1}{d} \left( \trace(M) + \sum_{t=1}^n \norm{x_t}_2^2 \right) \right)^d.
\end{align*}
\end{proof}

\begin{lemma} \label{lem:selfNorm}
Let $M_n = M + \sum_{t=1}^n \sum_{k=1}^{K_t} x_{t,k} x_{t,k}^{\top}$, where $M \in \RR^{d \times d}$ is a strictly positive definite matrix and $x_{t,k} \in \RR^d$ is a $d$-dimensional column vector. If $\sum_{k=1}^{K_t} \norm{x_{t,k}}_{M_{t-1}^{-1}}^2 \leq 1$, then
\begin{equation}
\sum_{t=1}^n \sum_{k=1}^{K_t} \norm{x_{t,k}}_{M_{t-1}^{-1}}^2 \leq 2\log\frac{\det(M_n)}{\det(M)}.
\end{equation}
Furthermore, if $\norm{x_{t,k}}_2 \leq L, K_t \leq K, \forall t,k$, and $M = \lambda I, \lambda \geq KL^2$, then
\begin{equation}
\label{eq:selfNormSum}
\sum_{t=1}^n \sum_{k=1}^{K_t} \norm{x_{t,k}}_{M_{t-1}^{-1}} \leq \sqrt{2dnK \log\left(1 + \frac{nKL^2}{\lambda d}\right)}.
\end{equation}
\end{lemma}
\begin{proof}
\begin{align*}
\det(M_n) &= \det(M_{n-1}) \det\left(I + M_{n-1}^{-1/2} \left(\sum_{k=1}^{K_n}x_{n,k} x_{n,k}^{\top} \right) M_{n-1}^{-1/2} \right)\\
&\overset{(a)}{\geq} \det(M_{n-1}) \left(1 + \sum_{k=1}^{K_n} \norm{x_{t,k}}_{M_{n-1}^{-1}}^2 \right) \geq \det(M) \prod_{t=1}^n \left(1 + \sum_{k=1}^{K_t} \norm{x_{t,k}}_{M_{t-1}^{-1}}^2 \right),
\end{align*}
where (a) is by Lemma A.3 of \cite{li2016contextual}. Then
$$
\sum_{t=1}^n \sum_{k=1}^{K_t} \norm{x_{t,k}}_{M_{t-1}^{-1}}^2 \overset{(b)}{\le} \sum_{t=1}^n 2 \log\left(1 + \sum_{k=1}^{K_t} \norm{x_{t,k}}_{M_{t-1}^{-1}}^2 \right) \leq 2\log\frac{\det(M_n)}{\det(M)},
$$
in which (b) is due to $2\log(1+u) \geq u$ for $u\in[0,1]$.

If $\norm{x_{t,k}}_2 \leq L, K_t \leq K, \forall t,k$, and $M = \lambda I, \lambda \geq KL^2$, then
\begin{align*}
\sum_{t=1}^n \sum_{k=1}^{K_t} \norm{x_{t,k}}_{M_{t-1}^{-1}} \le \sqrt{nK} \sqrt{ \sum_{t=1}^n \sum_{k=1}^{K_t} \norm{x_{t,k}}_{M_{t-1}^{-1}}^2 } \le \sqrt{2nK\log\frac{\det(M_n)}{\det(\lambda I)}} \le \sqrt{2dnK \log\left(1 + \frac{nKL^2}{\lambda d}\right)}.
\end{align*}
\end{proof}

\begin{lemma} \label{lem:lambdaMinOfM}
Let $\bx_t, t \geq 1$ be generated sequentially from a random distribution $\bx \in \RR^d$ such that $\norm{\bx}_2 \leq L$ and $\EE[\bx \bx^{\top}]$ is full rank with minimal eigenvalue $\lambda_x > 0$. Let $\bS_t = \sum_{s=1}^{t} \bx_s \bx_s^{\top}$. Let
\begin{equation}
A(\delta) = \log\frac{(tL^4 + 1)(tL^4 + 3)d}{\delta}.
\end{equation}
Then for any $\delta > 0$,
$$
\lambda_{\min}(\bS_t) \geq \left( t\lambda_x - \frac{L^2}{3}\sqrt{ 18tA(\delta) + A(\delta)^2 } - \frac{L^2}{3} A(\delta) \right)_+,
$$
holds for all $t \geq 0$ with probability at least $1-\delta$. Furthermore, if $L = 1, 0 < \delta \le \frac{1}{8}$, then the event
$$
\lambda_{\min}(\bS_t) \geq t\lambda_x/2, ~~\text{ for all   } t \ge \frac{256}{\lambda_x^2}\log\frac{128d}{\lambda_x^2 \delta}
$$
holds with probability at least $1-\delta$.
\end{lemma}
\begin{proof}
Let $\EE_{t}[\cdot] = \EE[\cdot | \bx_1, ..., \bx_t]$ and then
\begin{align*}
\bX_t &= \EE_{t-1} [\bx_t \bx_t^{\top}] - \bx_t \bx_t^{\top} = \EE[\bx \bx^{\top}] - \bx_t \bx_t^{\top},\\
\bY_t &= \sum_{s=1}^{t} \bX_s = t \cdot \EE[\bx \bx^{\top}] - \sum_{s=1}^{t} \bx_s \bx_s^{\top} = t \cdot \EE[\bx \bx^{\top}] - \bS_t.
\end{align*}
Then $\{\bY_t : t = 0, 1, 2, . . . \}$ is a matrix martingale whose values are self-adjoint matrices with dimension d and $\{\bX_s: s = 1,2,...\}$ is the difference sequence. Notice that $\lambda_{\max}(\bX_t) \leq L^2$. Then by Freedman's inequality for matrix martingales, Theorem 1.2 of \cite{tropp2011freedman}, for all $a \geq 0$ and $b>0$,
$$
\PP\left( \exists t\geq 0: \lambda_{\max}(\bY_t) \geq a \text{ and } \norm{W_t} \leq b \right) \leq d \cdot \exp\left( - \frac{a^2/2}{b + aL^2/3} \right),
$$
where
$$
W_t = \sum_{s=1}^{t} \EE_{s-1}[\bX_s^2] = \sum_{s=1}^{t} \left( \EE_{s-1}[(\bx_s \bx_s^{\top})^2] -  \EE[\bx \bx^{\top}]^2\right) \leq t \cdot \left( L^2 \EE[\bx \bx^{\top}] -  \EE[\bx \bx^{\top}]^2\right)
$$
and $\norm{\cdot}$ denotes the {\it spectral norm}. For any $a \geq \frac{1}{3} AL^2 + \sqrt{\frac{1}{9}A^2L^4 +2Ab } =: f(A,b)$,
$$
\frac{a^2/2}{b + aL^2/3} \geq A.
$$
Let $A(r,\delta) = \log\frac{(r+1)(r+3)}{\delta}$. Then
\begin{align*}
&\PP\left[ \exists t: \lambda_{\min}(\bS_t) \leq t\lambda_x - f(A(tL^4, \delta), tL^4) \right] \\
\le &\PP\left[\exists t: \lambda_{\min}(\bS_t) \leq t\lambda_x - f(A(\norm{W_t}, \delta), \norm{W_t})\right] \numberthis \label{eq:lemB3_1}\\
\le &\sum_{r=0}^{\infty} \PP\left[\exists t: \lambda_{\min}(\bS_t) \leq t\lambda_x - f(A(\norm{W_t}, \delta), \norm{W_t}), r-1 < \norm{W_t} \leq r \right]\\
\le &\sum_{r=0}^{\infty} \PP\left[\exists t: \lambda_{\min}(\bS_t) \leq t\lambda_x - f(A(r, \delta), r), \norm{W_t} \leq r \right] \\
\le &\sum_{r=0}^{\infty} \PP\left[\exists t: \lambda_{\max}(\bY_t) \geq f(A(r, \delta), r), \norm{W_t} \le r \right] \numberthis \label{eq:lemB3_2}\\
\le &d\sum_{r=0}^{\infty} \exp\left(- A(r, \delta) \right) = d\sum_{r=0}^{\infty} \frac{\delta}{(r+1)(r+3)} \leq d \cdot \delta,
\end{align*}
where \eqref{eq:lemB3_1} is because $A$ is increasing in $r$, $f$ is increasing in $A, b$, and
$$
\norm{W_t} = t \cdot \lambda_{\max}\left( L^2 \EE[\bx \bx^{\top}] -  \EE[\bx \bx^{\top}]^2\right) \leq t L^4,
$$
and \eqref{eq:lemB3_2} is by
$$
\lambda_{\max}(\bY_t) = t \lambda_{\max}(\EE[\bx \bx^{\top}]) - \lambda_{\min}(\bS_t) \geq t\lambda_x - \lambda_{\min}(\bS_t).
$$
The result is obtained by taking $\delta$ by $\delta/d$.

For the second part with $L=1$,
\begin{align*}
&t \geq \frac{256}{\lambda_x^2}\log\frac{128d}{\lambda_x^2 \delta}  \overset{(*)}{\Longrightarrow} t \geq \frac{128}{\lambda_x^2} \log\frac{td}{\delta}\\
\Longrightarrow &t\lambda_x/2 \geq \frac{2}{3}\sqrt{36 t \cdot 2\log\frac{td}{\delta} } \overset{(**)}{\ge} \frac{2}{3}\sqrt{18t \cdot 2\log\frac{td}{\delta} + \left(2\log\frac{td}{\delta} \right)^2} \overset{(***)}{\ge} \frac{1}{3} \sqrt{18tA(\delta) + A(\delta)^2} + \frac{1}{3}A(\delta)\\
\Longrightarrow &\lambda_{\min}(\bS_t) \geq t\lambda_x/2,
\end{align*}
where (*) is by the following Lemma \ref{lem:tGeqLogt}, (**) is by $\lambda_x \leq 1$ and $128/\lambda_x^2 \geq 2/18$, (***) is by $\delta \le 1/8 \le \frac{t^2d}{(t+1)(t+3)}$.
\end{proof}

\begin{lemma} \label{lem:sumBerGeq}
Let $x_1, x_2, \ldots, x_n$ be independent Bernoulli random variables with mean $0<p\le\frac{1}{2}$. Let $\delta > 0, B > 0$. Then with probability at least $1-\delta$,
\begin{equation}
\sum_{s=1}^t x_s \geq B, ~~\forall t \geq \frac{16}{p}\log\frac{n}{\delta} + \frac{4B}{p}.
\end{equation}
\end{lemma}
\begin{proof}
Note $p-x_s$ has mean zero, $|p-x_s| \leq 1-p<1$ and $\EE[(p-x_s)^2] = p(1-p)<p$. Then by Bernstein's inequality with fixed $t$,
$$
\PP \left( tp  - \sum_{s=1}^t x_s > a \right) \leq \exp \left( - \frac{a^2/2}{tp + a/3} \right).
$$
Fix $t \ge \frac{16}{p}\log\frac{1}{\delta} + \frac{4B}{p}$, then $tp-B \ge 4\sqrt{tp\log\frac{1}{\delta}}$. Thus
$$
\PP\left( \sum_{s=1}^t x_s < B \right) = \PP\left( tp - \sum_{s=1}^t x_s >tp-B \right) \le \PP\left( tp - \sum_{s=1}^t x_s > 4\sqrt{tp\log\frac{1}{\delta}}\right) \le \delta.
$$
Replace $\delta$ by $\delta/n$ to reach the result.
\end{proof}

\begin{lemma} \label{lem:tGeqLogt}
If $a>0, b>0, ab\geq e$, then for all $t \geq 2a\log(ab)$,
\begin{equation}
\label{eq:tGeqLogt}
t \geq a\log(bt).
\end{equation}
\end{lemma}
\begin{proof}
The LHS of \eqref{eq:tGeqLogt} increases faster than the RHS. So it suffices to prove \eqref{eq:tGeqLogt} holds for $t = 2a\log(ab)$, or equivalently
$$
2a\log(ab) \geq a\log(2ab\log(ab)) = a\log(ab) + a\log(2\log(ab)),
$$
which can be easily derived by observing that $ab \geq 2\log(ab)$.
\end{proof}

\begin{lemma} \label{lem:tForLeGamma}
When $T \ge \frac{512d}{\gamma^2\lambda_x}\ln\frac{4u}{\delta}$,
\begin{align*}
\frac{\sqrt{d\ln\left(1 + \frac{T}{\lambda d}\right) + 2\ln\frac{4u}{\delta}} + \sqrt{\lambda}}{\sqrt{\lambda + T \lambda_x /8}} \le \frac{\gamma}{2}\,.
\end{align*}
\end{lemma}
\begin{proof}
Assume $\lambda \le d\ln(1 + \frac{T}{\lambda d}) + 2\ln\frac{4u}{\delta}$ (which is typically held). It suffices to prove that
\begin{align*}
\frac{d\ln\left(1 + \frac{T}{\lambda d}\right) + 2\ln\frac{4u}{\delta}}{T \lambda_x /8} \le \frac{\gamma^2}{16}
\end{align*}
or it suffices to prove that
\begin{align*}
\frac{d\ln\left(1 + \frac{T}{\lambda d}\right)}{T \lambda_x /8} \le \frac{\gamma^2}{32} \qquad \text{and} \qquad \frac{2\ln\frac{4u}{\delta}}{T \lambda_x /8} \le \frac{\gamma^2}{32}\,.
\end{align*}
The first one requires $T \ge \frac{512d}{\gamma^2 \lambda_x}\ln\frac{512}{\gamma^2\lambda_x\lambda}$ by Lemma \ref{lem:tGeqLogt} and the second one requires $T \ge \frac{512}{\gamma^2 \lambda_x}\ln\frac{4u}{\delta}$. The condition on $T$ would satisfy both if $\delta \le \frac{u\gamma^2\lambda_x \lambda}{128}$ which is typically held.

\end{proof}

\end{document}